\documentclass[sigconf]{acmart}

\usepackage{wrapfig}
\usepackage{graphicx}
\usepackage{subcaption}
\usepackage{printlen}
\usepackage{float}
\uselengthunit{cm}

\newcommand{\nodesize}{0.028\textwidth}
\usepackage{amsmath}
\newcommand{\squishlist}{
	\begin{list}{$\bullet$}
		{ \setlength{\itemsep}{1pt}
			\setlength{\parsep}{1pt}
			\setlength{\topsep}{2.5pt}
			\setlength{\partopsep}{0.5pt}
			\setlength{\leftmargin}{1em}
			\setlength{\labelwidth}{1em}
			\setlength{\labelsep}{0.6em}
		}
	}
	\newcommand{\squishend}{
	\end{list}
}

\usepackage{tikz}
\usetikzlibrary{arrows.meta, positioning}
\usepackage{xcolor}
\usepackage[linesnumbered,ruled,vlined]{algorithm2e}
\usepackage{enumitem}
\usepackage{multirow}
\usepackage{colortbl}
\usepackage{pifont}

\newcommand{\first}[1]{\cellcolor[rgb]{ .549,  .71,  .976}\textbf{#1}}
\newcommand{\second}[1]{\cellcolor[rgb]{ .702,  .808,  .984}{#1}}
\newcommand{\third}[1]{\cellcolor[rgb]{ .851,  .906,  .992}{#1}}
\definecolor{mygreen}{RGB}{130, 172, 211}
\definecolor{myblue}{RGB}{222, 235, 247}

\definecolor{lightblue}{RGB}{102, 178, 255}

\AtBeginDocument{%
	}

\setcopyright{acmlicensed}
\copyrightyear{2018}
\acmYear{2018}
\acmDOI{XXXXXXX.XXXXXXX}

\acmConference[Conference acronym 'XX]{Make sure to enter the correct
	conference title from your rights confirmation emai}{June 03--05,
	2018}{Woodstock, NY}
\acmBooktitle{Woodstock '18: ACM Symposium on Neural Gaze Detection,
 June 03--05, 2018, Woodstock, NY}
\acmISBN{978-1-4503-XXXX-X/18/06}

\begin{document}
	\title{Retrofitting Temporal Graph Neural Networks with Transformer} 
	
	\author{Qiang Huang}
	\email{2023102110034@whu.edu.cn}
	\affiliation{%
		\institution{School of Computer Science,\\ Wuhan University}
		\country{China}
	}

        \author{Xiao Yan}
	\email{yanxiaosunny@gmail.com}
	\affiliation{%
		\institution{Centre for Perceptual and Interactive Intelligence (CPII)}
		\country{Hong Kong, China}
	}

        \author{Xin Wang}
	\email{2019302110391@whu.edu.cn}
	\affiliation{%
		\institution{School of Computer Science,\\ Wuhan University}
		\country{China}
	}

        \author{Susie Xi Rao}
	\email{raox@inf.ethz.ch}
	\affiliation{%
		\institution{ETH Zürich}
		\country{Switzerland}
	}

        \author{Zhichao Han}
	\email{zhihan@ebay.com}
	\affiliation{%
		\institution{eBay}
		\country{China}
	}

        \author{Fangcheng Fu}
	\email{ccchengff@pku.edu.cn}
	\affiliation{%
		\institution{Peking University}
		\country{China}
	}

        \author{Wentao Zhang}
	\email{wentao.zhang@pku.edu.cn}
	\affiliation{%
		\institution{Peking University}
		\country{China}
	}
	
	\author{Jiawei Jiang}
	\authornote{Jiawei Jiang is the corresponding author.}
	\email{jiawei.jiang@whu.edu.cn}
	\affiliation{%
		\institution{School of Computer Science,\\ Wuhan University}
		\country{China}
	}

	%\affiliation{%
	%  \institution{Inria Paris-Rocquencourt}
	%\affiliation{%
	% \institution{Rajiv Gandhi University}
	%\affiliation{%
	%  \institution{Tsinghua University}
	%\affiliation{%
	%  \institution{Palmer Research Laboratories}
	%\affiliation{%
	%  \institution{The Th{\o}rv{\"a}ld Group}
	%\email{jsmith@affiliation.org}
	%\affiliation{%
	%  \institution{The Kumquat Consortium}
	%  \city{New York}
	%\email{jpkumquat@consortium.net}
	
	\renewcommand{\shortauthors}{Trovato et al.}
	
	\begin{abstract}
        Temporal graph neural networks (TGNNs) outperform regular GNNs by incorporating time information into graph-based operations. However, TGNNs adopt specialized models (e.g., TGN, TGAT, and APAN ) and require tailored training frameworks (e.g., TGL and ETC). In this paper, we propose TF-TGN, which uses Transformer decoder as the backbone model for TGNN to enjoy Transformer's codebase for efficient training. In particular, Transformer achieves tremendous success for language modeling, and thus the community developed high-performance kernels (e.g., flash-attention and memory-efficient attention) and efficient distributed training schemes (e.g., PyTorch FSDP, DeepSpeed, and Megatron-LM). We observe that TGNN resembles language modeling, i.e.,  the message aggregation operation between chronologically occurring nodes and their temporal neighbors in TGNNs can be structured as sequence modeling. Beside this similarity, we also incorporate a series of algorithm designs including suffix infilling, temporal graph attention with self-loop, and causal masking self-attention to make TF-TGN work. During training, existing systems are slow in transforming the graph topology and conducting graph sampling. As such, we propose methods to parallelize the CSR format conversion and graph sampling. We also adapt Transformer codebase to train TF-TGN efficiently with multiple GPUs. We experiment with 9 graphs and compare with 2 state-of-the-art TGNN training frameworks. The results show that TF-TGN can accelerate training by over 2.20$\times$ while providing comparable  or even superior accuracy  to existing SOTA TGNNs. TF-TGN is available at \url{https://github.com/qianghuangwhu/TF-TGN}.

	\end{abstract}
	
	
	% 	\includegraphics[width=\textwidth]{sampleteaser}
	% 	\caption{Seattle Mariners at Spring Training, 2010.}
	% 	\Description{Enjoying the baseball game from the third-base
	% 		seats. Ichiro Suzuki preparing to bat.}
	% 	\label{fig:teaser}

	\received{20 February 2007}
	\received[revised]{12 March 2009}
	\received[accepted]{5 June 2009}
	
	\maketitle
	
	\section{Introduction}

    Dynamic graphs are networks whose structures are continuously changing and evolving over time and are widely employed to model interactions among entities across various domains, including social networks, e-commerce, and biological networks~\cite{wasserman1994social, kumar2019predicting,rossi2020temporal,lu2022bright, wu2022bridgedpi}. 
    Recently, temporal graph neural networks (TGNNs) have been developed to model the temporal, structural, and evolution of dynamic graphs. 
    % TGNNs incorporate graph-based operations, such as graph convolution~\cite{kipf2016semi}, graph attention~\cite{vaswani2017attention, velivckovic2017graph}, message passing~\cite{gilmer2017neural,gilmer2020message} and motifs~\cite{paranjape2017motifs}, with temporal information to model the dynamic evolution of node features. 
    TGNNs have been demonstrated to outperform static GNNs by incorporating temporal information into graph-based operations across tasks such as dynamic link prediction and dynamic node classification~\cite{bai2020temporal, trivedi2019dyrep, kumar2019predicting, xu2020inductive, rossi2020temporal}.
    Popular TGNN models include TGN~\cite{rossi2020temporal}, TGAT~\cite{xu2020inductive}, APAN~\cite{wang2021apan}, and CAWN~\cite{wang2021inductive}
    are proposed to  accurately simulate the  evolution of temporal graphs. These models heavily rely on  graph-based operations such as memory-based aggregation and update operations, classical graph attention, and motifs. Several efficient TGNN training systems including TGL~\cite{zhou2022tgl}, and ETC~\cite{gao2024etc} are also designed to train TGNNs efficiently.

\begin{figure}[!tb]
    \centering
        \includegraphics[width=0.6\linewidth]{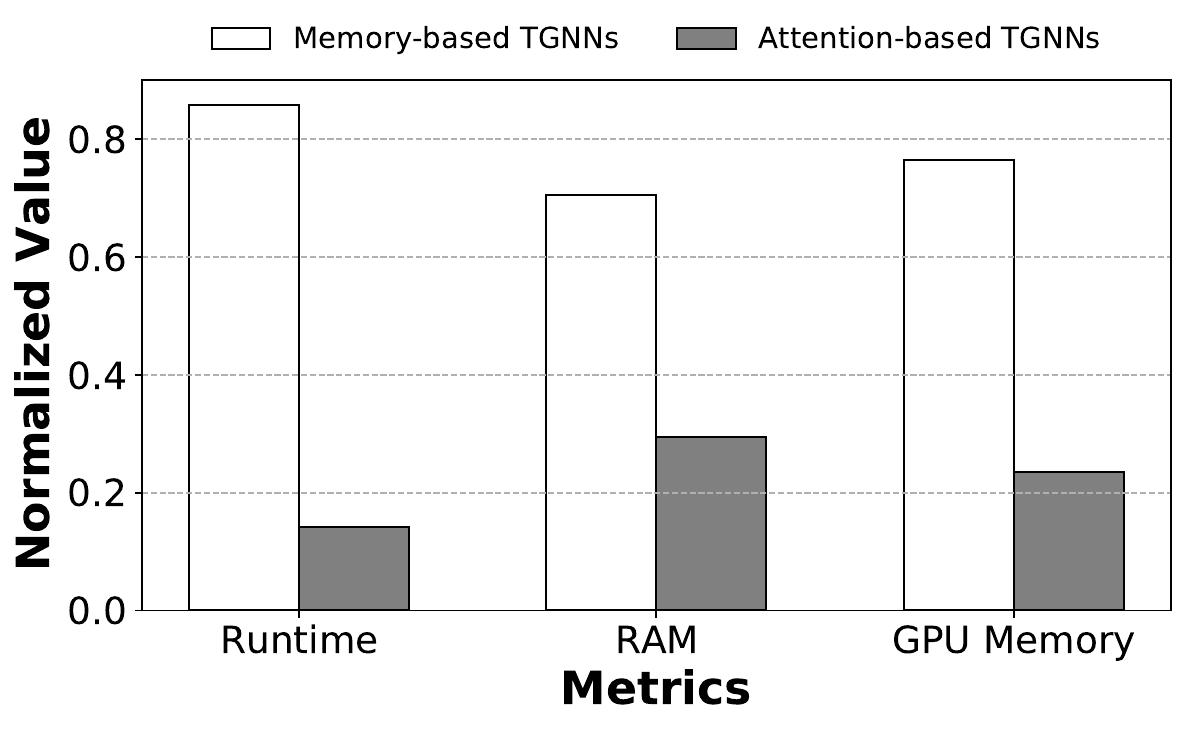}
        \caption{Comparison of normalized values for GPU memory, RAM, and runtime between memory-based and attention-based TGNNs.}
        \label{fig:memory_attention}
\end{figure}

 Transformer decoder is a model architecture that uses causal masking self-attention for sequence modeling and has been very successful for natural language processing (NLP). As such, the community developed a rich codebase for the efficient execution of Transformers, ranging from attention kernels such as flash-attention~\cite{dao2022flashattention,dao2023flashattention}, and memory-efficient attention~\cite{rabe2021self}, as well as distributed training schemes such as PyTorch FSDP~\cite{paszke2019pytorch}, DeepSpeed~\cite{rasley2020deepspeed}, and Megatron-LM~\cite{shoeybi2019megatron}. Furthermore, recent literatures have revealed that the attention-based TGNNs are more efficient than memory-based TGNNs in modeling temporal graphs~\cite{xu2020inductive,rossi2020temporal,poursafaei2022towards,yu2023towards,huang2024benchtemp}.  In Figure~\ref{fig:memory_attention}, we experiment the graph kernel of memory-based TGN~\cite{rossi2020temporal} and  attention-based TGAT~\cite{xu2020inductive} under similar configurations. The results show that attention kernel has smaller costs in all aspects,  suggesting that casting TGNN to Transformer benefits efficiency.  
    
    To realize the idea, we design the Transformer-based TGNN (TF-TGN for short) by tackling two main technical challenges. 
    
    \ding{182} \textit{How to adapt Transformer to model TGNN}? 
    Transformer has become the de-facto standard for sequence modeling, especially in NLP.
    In a temporal graph, nodes are associated with timestamps and occur chronologically, allowing temporal neighbors to be arranged in sequence.
    The message aggregation operation in TGNNs between nodes and their temporal neighbors can be structured as sequence modeling by suffix infilling the current node at the end of the temporal neighbors, combined with the temporal graph attention with self-loop operation. 
    
    \ding{183} \textit{How to train Transformer-based TGNN efficiently}? 
    Based on sequence modeling, high-performance kernels in Transformer, such as flash-attention and memory-efficient attention, along with efficient distributed training schemes like PyTorch FSDP and DeepSpeed, can be directly applied to TF-TGN for efficient TGNN training.
    Besides model computation, we also observe that graph format conversion and graph sampling are slow in existing systems. 
    In particular, graph sampling is used to determine the edges for computation, and the the graph needs to be converted into temporal compressed sparse row representation (T-CSR) for time-based sampling. Format conversion can constitute more than 30\% of the training time. Therefore, we proposed a parallel sampling strategy that leverages concurrent threads and atomic access to accelerate the CSR conversion and sampling.
    
    To evaluate TF-TGNN, we conduct extensive experiments on 9 datasets and compare with 2 state-of-the-art TGNN training frameworks. The results show that TF-TGNN matches the accuracy of existing TGNNs. Regarding end-to-end training time, TF-TGN accelerates TGNN frameworks by 2.20$\times$ on average across the datasets and 10.31$\times$ at the maximum. For graph format conversion and graph sampling, TF-TGN yields a speedup of up to 1466.45$\times$ and 16.9$\times$, respectively. 
    
    % Moreover, TF-TGN also scales to training with 8 GPUs.      
    
    To summarize, we make the following contributions. 
    \squishlist
    \item We observe that the message aggregation operation in TGNNs resembles sequence modeling. Motivated as such, we propose TF-TGN to adapt TGNNs to the Transformer decoder for effectively modeling temporal graphs while leveraging its highly optimized codebase.
    
    \item We propose a series of algorithm designs including suffix infilling, temporal graph attention with self-loop, and causal masking self-attention to run TGNNs with Transformer.
    
    \item We adjust the Transformer's codebase for TGNN training, incorporating flash-attention, memory-efficient attention, and distributed training schemes, while parallelizing graph CSR format conversion and sampling for efficiency.
    \item  Extensive experiments are conducted on 9 real-world temporal graphs with up to  billions of edges to demonstrate  the effectiveness of TF-TGN in speedup and prediction accuracy.  
    
    \squishend

\section{Preliminaries}

\noindent\textbf{Dynamic Graphs.}
% \textbf{Dynamic Graphs.}
 A dynamic graph $G(\mathcal{V}(t), \mathcal{E}(t), \phi, \psi)$ at time $t$  can be represented as an ordered sequence of temporal interactions~$S_{t}=\{s_{1}, \dots, s_{i}, \dots,  s_{n}\}$. The $i$-th interaction $s_{i}=(u_{i}, v_{i}, t_{i}, \mathbf{e}_{i})$ happens at time $t_{i}$ between the node $u_{i}$ and the node $v_{i}$ with edge feature $\mathbf{e}_{i}$~ ~\cite{rossi2020temporal}. $\phi: \mathcal{V} \rightarrow \mathcal{A}$ maps nodes of each temporal edge to their types, while $\psi: \mathcal{E} \rightarrow \mathcal{R}$ maps temporal edge to its type. $\mathcal{A}$ and $\mathcal{R}$ are the sets of node types and edge types, respectively. Table \ref{tab:notation} lists the notations used in this paper.

% \textbf{Temporal Graph Neural Networks.}
% Temporal graph neural networks have been proposed to model the dynamic evolution of temporal graphs~\cite{kim2018review, trivedi2019dyrep, kumar2019predicting, xu2020inductive, rossi2020temporal}. TGNNs can be divided into two categories: discrete-time dynamic graphs (DTDG) and continuos-time dynamic graphs (CTDG). DTDG models the temporal graph as a sequence of static graph snapshots, while CTDG models the temporal graph as a continuous-time graph ~\cite{rossi2020temporal}. 

% \noindent\textbf{Temporal Graph Attention.} 
\noindent\textbf{Temporal Graph Attention.}
\label{subsec:TGAT}
Temporal graph attention based on self-attention is a scalable module that can effectively and efficiently aggregate the features of temporal neighborhoods in dynamic graphs~\cite{xu2020inductive,huang2024benchtemp}. Given a node $v(t) \in \mathcal{V}(t)$ and its $k$ temporal neighbors $\mathcal{N}_{v}(t) = \{v_{1}(t_{1}), \dots, v_k(t_{k})\}$ sampled by a temporal neighbor sampler at time $t$ and the timestamps of the temporal neighbors are $\{t_{1}, \dots, t_{k}\}$, respectively. The hidden state of node $v(t)$ is $\mathbf{h}_{v}(t)\in \mathbb{R}^{1\times~d_{v}}$ and the hidden states of its temporal neighbors are $[\mathbf{h}_{v_1}(t_1), \dots, \mathbf{h}_{v_k}(t_k)]^{\top} \in \mathbb{R}^{k\times~d_{v}}$. The temporal edge features between the node $v(t)$ and its temporal neighbors are $[\mathbf{e}_{1}(t_1), \dots, \mathbf{e}_{k}(t_k)]^{\top} \in \mathbb{R}^{k\times~d_{e}}$. The time encoding function $\Phi$ encodes the time intervals between the node $v(t)$ and its temporal neighbors $\Delta{\mathbf{t}}=[t-t_1, \dots, t-t_k]^{\top}$ into a time embedding matrix $[\Phi([\Delta{\mathbf{t}}]_{1}), \dots, \Phi([\Delta{\mathbf{t}}]_{k})]^{\top} \in \mathbb{R}^{k\times~d_t}$. Let $\mathbf{Z}_{v}(t)=[\mathbf{h}_{v}(t)\oplus\Phi(0), \mathbf{h}_{v_1}(t_1)\oplus\Phi([\Delta{\mathbf{t}}]_{1})\oplus~\mathbf{e}_{1}(t_1),\dots,\mathbf{h}_{v_k}(t_k)\oplus\Phi([\Delta{\mathbf{t}}]_{k})\oplus~\mathbf{e}_{k}(t_k)]^{\top}$, where $\oplus$ denotes the concatenation operation or sum operation.
The query $\mathbf{q}(t)$, key $\mathbf{K}(t)$, and value $\mathbf{V}(t)$ can be formulated as 
% \begin{equation}
%     \centering
%     \label{eq:qkv}
%     \begin{aligned}
%         \centering
%         \mathbf{q}(t) & = [\mathbf{Z}_{v}(t)]_{0}\mathbf{W}_{Q}, \\ 
%         \mathbf{K}(t) & = [\mathbf{Z}_{v}(t)]_{1:k}\mathbf{W}_{K},\\
%         \mathbf{V}(t)& =[\mathbf{Z}_{v}(t)]_{1:k}\mathbf{W}_{V},
%     \end{aligned}
% \end{equation}
\begin{equation}
    \centering
    \label{eq:qkv}
    \small
    \begin{aligned}
        \centering
        \mathbf{q}(t)  = [\mathbf{Z}_{v}(t)]_{0}\mathbf{W}_{q}, 
        \mathbf{K}(t)  = [\mathbf{Z}_{v}(t)]_{1:k}\mathbf{W}_{K},
        \mathbf{V}(t) =[\mathbf{Z}_{v}(t)]_{1:k}\mathbf{W}_{V},
    \end{aligned}
\end{equation}
% \begin{equation*}
%     \centering
%         \mathbf{Q}(t)  = [\mathbf{Z}_{v}(t)]_{0}\mathbf{W}_{Q}; \mathbf{K}(t)  = [\mathbf{Z}_{v}(t)]_{1:n}\mathbf{W}_{K}; \mathbf{V}(t) =[\mathbf{Z}_{v}(t)]_{1:n}\mathbf{W}_{V}.
% \end{equation*}
where $\mathbf{W}_{q}\in \mathbb{R}^{(d_v+d_t) \times d_{k}}, \mathbf{W}_{K}, \mathbf{W}_{V} \in \mathbb{R}^{d \times d_{k}}$  denote the learnable parameters, $d=d_v+d_e+d_t$. Then, the updated hidden state of node $v$ at time $t$ can be calculated by aggregating the features associated with its temporal neighbors using a self-attention mechanism as follows:
\begin{equation}
    \small
    \tilde{\mathbf{h}}_v(t) = \text{softmax}(\frac{\mathbf{q}(t)\mathbf{K}(t)^{\top}}{\sqrt{d_k}})\mathbf{V}(t).
\end{equation}
\begin{figure*}[!tb]
    \centering
    \begin{subfigure}[b]{.22\textwidth}
        \centering
        \begin{tikzpicture}[
            every node/.style={circle, draw, align=center, minimum size=\nodesize, font=\sffamily\bfseries\large, text=black}, 
            level distance=1.3cm,
            level 1/.style={sibling distance=1cm},
            edge from parent/.style={draw, line width=0.3mm, font=\sffamily\bfseries\large}]
        
            \node[fill=mygreen,dashed] {7}
            child {node[fill=myblue] {1} edge from parent node [midway, left, yshift=-1mm, draw=none, text=black] {$t_1$}}
            child {node[fill=myblue] {2} edge from parent node [midway, xshift=-2mm, yshift=-1mm, draw=none, text=black] {$t_2$}}
            child {node[fill=myblue] {5} edge from parent node [midway, right, yshift=-1mm, draw=none, text=black] {$t_3$}};
        \end{tikzpicture}
        \caption{A node and its temporal neighbors.}
    \end{subfigure}
    \hfil
    \begin{subfigure}[b]{.23\textwidth}
        \centering
        \begin{tikzpicture}[
            every node/.style={circle, draw, align=center, minimum size=\nodesize, font=\sffamily\bfseries\large, text=black}
        ]
            % Draw the horizontal line
            \draw[thick,->] (-0.3,0) -- (3.3,0);
            \foreach \x in {0,1,2} {
                \draw (\x, 0.1) -- (\x, -0.1) node[below, draw=none, text=black] {$t_{\number\numexpr\x+1\relax}$}; % Tick mark with label below
            }
            \draw (3, 0.1) -- (3, -0.1) node[below, draw=none, text=black] {$t_{ ~}$};
            
            % Nodes
            \node[draw, circle, fill=myblue] (n1) at (0,0.7) {1};
            \node[draw, circle, fill=myblue] (n3) at (1,0.7) {2};
            \node[draw, circle, fill=myblue] (n5) at (2,0.7) {5};
            \node[draw, circle, fill=mygreen,dashed] (n7) at (3,0.7) {7};
        
            % Labels
            % \node[draw=none, text=black] at (0.5,-0.5) {$t_1$};
            % \node[draw=none, text=black] at (1.5,-0.5) {$t_2$};
            % \node[draw=none, text=black] at (2.5,-0.5) {$t_3$};
            % \node[draw=none, text=black] at (3.5,-0.5) {$t$};
        \end{tikzpicture}
        \caption{Suffix infilling as a chronological sequence. \label{fig:suffix_infilling}}
    \end{subfigure}
    \hfil
    \begin{subfigure}[b]{.23\textwidth}
        \centering
        \begin{tikzpicture}[
            every node/.style={circle, draw, align=center, minimum size=\nodesize, font=\sffamily\bfseries\large, text=black}, 
            level distance=1.3cm,
            level 1/.style={sibling distance=1cm},
            edge from parent/.style={draw, line width=0.3mm, font=\sffamily\bfseries\large}]
        
            \node[fill=mygreen,dashed] (node7) {7}
            child {node[fill=myblue] {1} edge from parent node [midway, left, yshift=-1mm, draw=none, text=black] {$\alpha_1$}}
            child {node[fill=myblue] {2} edge from parent node [midway, xshift=-2mm, yshift=-1mm, draw=none, text=black] {$\alpha_2$}}
            child {node[fill=myblue] {5} edge from parent node [midway, right, yshift=-1mm, draw=none, text=black] {$\alpha_3$}};
            edge [loop above] node [xshift=2mm, yshift=5mm, text=black];
    
            \draw[->, >=stealth, thick, black]
    (node7.north) to[out=75, in=40, looseness=3] node[above right, midway, draw=none, text=black] {$\alpha$} (node7.east);
        \end{tikzpicture}
        \caption{Temporal graph attention with self-loop.\label{fig:attention_loop}}
    \end{subfigure}
    \hfil
    \begin{subfigure}[b]{.23\textwidth}
        \centering
        \begin{tikzpicture}[
            every node/.style={circle, draw, align=center, minimum size=\nodesize, font=\sffamily\bfseries\large, text=black}
        ]
            % Nodes
            \node[circle, draw, fill=myblue] (node1) at (0,0) {1};
            \node[circle, draw, fill=myblue] (node2) at (1,0) {2};
            \node[circle, draw, fill=myblue] (node5) at (2,0) {5};
            \node[circle, draw, fill=mygreen,dashed] (node7) at (3,0) {7};
            
            \node[circle, draw, fill=myblue] (node1b) at (0,1.2) {1};
            \node[circle, draw, fill=myblue] (node2b) at (1,1.2) {2};
            \node[circle, draw, fill=myblue] (node5b) at (2,1.2) {5};
            \node[circle, draw, fill=mygreen,dashed] (node7b) at (3,1.2) {7};
        
            % Edges
            \draw[color=black, line width=0.2mm] (node1b) -- (node1);
    
            \draw[color=black, line width=0.2mm] (node2b) -- (node1);
            \draw[color=black, line width=0.2mm] (node2b) -- (node2);
    
            \draw[color=black, line width=0.2mm] (node5b) -- (node1);
            \draw[color=black, line width=0.2mm] (node5b) -- (node2);
            \draw[color=black, line width=0.2mm] (node5b) -- (node5);
    
            \draw[color=black!60, line width=0.3mm] (node7b) -- (node1) node[midway, left, xshift=-1mm, draw=none, text=black] {$\alpha_1$};
            \draw[color=black!60, line width=0.3mm] (node7b) -- (node2) node[midway, left, draw=none, text=black] {$\alpha_2$};
            \draw[color=black!60, line width=0.3mm] (node7b) -- (node5) node[midway, left, xshift=1mm, draw=none, text=black] {$\alpha_3$};
            \draw[color=black!60, line width=0.3mm] (node7b) -- (node7) node[midway, left, xshift=1mm, draw=none, text=black] {$\alpha$};;
        \end{tikzpicture}
        \caption{Attention with causal masking.\label{fig:attention_causal}}
    \end{subfigure}
    \caption{
    (a) The neighbors of a node sampled by the temporal sampler; 
    (b) Suffix infilling the sampled temporal neighbors using the node itself as a sequence~$\mathcal{X}_u(t)$;
    (c) Temporal graph attention with self-loop;
    (d) Attention with causal masking of transfomer decoder on the suffix infilling sequence.}
    \end{figure*}
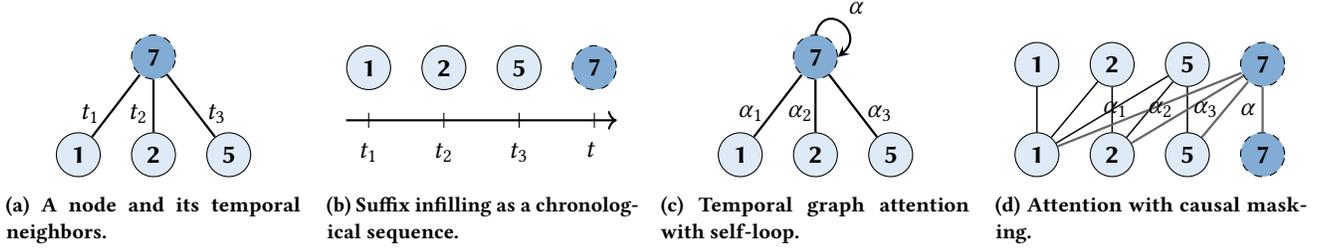
    
% \noindent\textbf{Autoregressive Sequence Modeling.}
\noindent\textbf{Autoregressive Sequence Modeling.}
For a $l$-length sequence $\mathcal{X}=\{x_1, \dots, x_l\}$, the autoregressive sequence modeling is to predict the next element $x_{i}$ based on the previous elements $\{x_1, \dots, x_{i-1}\}$ and can be formulated as the product of conditional probability distributions with a neural network parameterized by $\theta$~\cite{graves2013generating,radford2018improving, vaswani2017attention}: 
\begin{equation}
    \small
    P(\mathcal{X})=\prod_{i=1}^l P\left(x_i \mid x_1, \ldots, x_{i-1} ; \theta\right).
\end{equation}

The training objective of the neural network is to maximize the log-probability of the sequence data: 
\begin{equation}
    \small
    \mathcal{L}(\mathbf{x})=\sum_{i=1}^{l} \log P\left(x_i \mid x_{1}, \ldots, x_{i-1} ; \theta\right).
\end{equation}
\begin{table}[!tb]
    \small
    \centering
    \caption{Notations.}
    \label{tab:notation}
    \begin{tabular}{c|l}
        \hline
        Notation & Description \\
        \hline
        $G(\mathcal{V}(t), \mathcal{E}(t))$ & A dynamic graph with nodes $\mathcal{V}(t)$ and edges $\mathcal{E}(t)$ \\
        $s_{i}$ & $i$-th interaction of temporal interactions sequence $S$  \\
        $u_{i}$ and $v_{i}$  & The nodes associated with edge $s_{i}$ \\
        $\mathbf{h}_{v}$ & Hidden state of node $v \in \mathcal{V}_{t}$ \\
        $\mathbf{e}_{i}$ & The feature of temporal edge $(u_{i}, v_{i}) \in \mathcal{E}_{t}$ \\
        $\phi$ & Node type mapping function \\
        $\psi$ & Edge type mapping function \\
        $\mathcal{N}_{v}(t)$ & Temporal neighbors of node $v$ at time $t$\\
        $\Phi(\Delta{t})$ & Time encoding function of time interval $\vec{t}$ \\
        $\mathcal{X}$ & A sequence of elements \\
        \hline
    \end{tabular}
\end{table}
% \section{Methods}

% In this section, we introduce the distributed temporal graph attention network based on the  sequence modeling using the causal masking self-attention of the Transformer decoder. 

% First, we introduce a parallel sampling strategy to efficiently sample the temporal neighbors of each node. Then, we formulate the temporal message aggregation between nodes and their temporal neighbors of temporal graphs or static graphs as an autoregressive sequence modeling based on a suffix infilling operation and temporal graph attention with self-loop. Finally, we implement TF-TGN on top of flash-attention and memory-efficient attention mechanism in a distributed manner.

\section{Adapting to Transformer}
In this section, we introduce the straightforward yet innovative approach we developed to adapt the temporal message aggregation operation of TGNNs to the Transformer decoder architecture for aggregating temporal neighbor features.
We first introduce the suffix infilling operation to format the temporal neighbors of a node as a chronological sequence. Then, we propose a temporal attention with self-loop operation to consider the current node feature in attention. Finally, we introduce the causal masking self-attention of  Transformer decoder to model the temporal message aggregation between nodes and their temporal neighbors as sequence modeling.
 
\subsection{Suffix Infilling}
Infilling is a key operation in autoregressive sequence modeling, which is to predict the missing elements according to the surrounding elements~\cite{bavarian2022efficient, roziere2023code}.
For instance, in large language models (LLMs) for code, \textit{suffix-prefix-middle} infilling and \textit{prefix-suffix-middle} infilling are two common infilling operations to simulate the cursor movements on a IDE for the code generation and code completion tasks~\cite{bavarian2022efficient, roziere2023code}.

In a temporal graph, nodes are associated with timestamps and occur chronologically, allowing temporal neighbors to be arranged in sequence. \textit{To format a node $v(t)\in\mathcal{V}(t)$ at time $t$ and its temporal neighbors occurring before it as a chronological sequence}, we adopt a \textit{suffix infilling} operation to concatenate the the temporal neighbors $\overrightarrow{\mathcal{N}}_{v}(t) = (v_{1}, \dots, v_{k})$ sorted in ascending with the timestamps
$(t_{1}, \dots, t_{k})$
and the node $v$ itself at time $t$, $k = |\overrightarrow{\mathcal{N}}_{v}(t)|$.
 $\overrightarrow{\mathcal{N}}_{v}(t)$ can be easily and efficiently sampled by Algorithm \ref{alg:sampling} (as we will describe in Section \ref{subsec:sampling}).
The suffix infilling operation can be formulated as follows:
\begin{equation}
    \small
    \mathcal{X}_v(t) \leftarrow \overrightarrow{\mathcal{N}}_{v}(t) \| v(t),
\end{equation}
where $\|$ denotes the concatenation operation. Thus, $\mathcal{X}_u(t)$ is a sequence of elements as shown in Figure \ref{fig:suffix_infilling}:
\begin{equation}
    \small
    \mathcal{X}_v(t) = (v_{1}(t_1), \dots, v_{k}(t_k), v(t)),
\end{equation}
where $t_{1}<\dots<t_{k}<t$.
% \overset{\rightharpoonup}{a}

\subsection{Temporal Graph Attention with Self-loop}
In Equation \ref{eq:qkv}, the query $\mathbf{q}(t)$, key $\mathbf{K}(t)$, and value $\mathbf{V}(t)$ are calculated based on the hidden states of the node $v$ and its temporal neighbors, respectively, without considering the current node feature, i.e., the first element in $\mathbf{Z}_v(t)$. However, in sequence modeling with the Transformer decoder, the causal masking self-attention mechanism includes the current element.

To align the causal masking self-attention of the Transformer decoder, in which \textit{the attention  is allowed to attend to the previous elements and the current element.}
Therefore, based on the TGAT, we further proposed a variant of temporal graph attention with self-loop, which adds a self-loop edge between the node $v$ and itself as shown in Figure \ref{fig:attention_loop} to consider the current node feature in attention.
The temporal attention with self-loop can be formulated as follows:
\begin{equation}
    \centering
    \small
    \label{eq:qkv_selfloop}
    \begin{aligned}
        \centering
        \mathbf{q}(t) = [\mathbf{Z}_{v}(t)]_{0}\mathbf{W}_{Q}, 
        \mathbf{K}_{self}(t)  = [\mathbf{Z}_{v}(t)]_{0:k}\mathbf{W}_{K},
        \mathbf{V}_{self}(t) =[\mathbf{Z}_{v}(t)]_{0:k}\mathbf{W}_{V}.
    \end{aligned}
\end{equation}
% \begin{equation}
%     \centering
%     \label{eq:qkv_selfloop}
%     \begin{aligned}
%         \centering
%         \mathbf{q}(t) & = [\mathbf{Z}_{v}(t)]_{0}\mathbf{W}_{Q}, \\ 
%         \mathbf{K}_{self}(t) & = [\mathbf{Z}_{v}(t)]_{0:k}\mathbf{W}_{K},\\
%         \mathbf{V}_{self}(t)& =[\mathbf{Z}_{v}(t)]_{0:k}\mathbf{W}_{V}.
%     \end{aligned}
% \end{equation}
% 放置的位置需要改变
\subsection{Causal Masking Self-attention}
% \begin{figure}[!tb]
%     \centering
%     \begin{subfigure}[b]{.22\textwidth}
%         \includegraphics[width=\linewidth]{figs/batch_seqs.pdf}
%         \caption{Batch of suffix infilling temporal sequences.}
%     \end{subfigure}
%     \hfil
%     \begin{subfigure}[b]{.22\textwidth}
%         \includegraphics[width=\linewidth]{figs/causal_masking.pdf}
%         \caption{The causal masking for the batch of suffix infilling temporal sequences.}
%     \end{subfigure}
%     \caption{The causal masking for the batch of suffix infilling temporal sequences. Cells in gray are padding encodes and are masked in the self-attention mechanism. Long temporal sequences are truncated to the maximum neighbor size.} 
%     \label{fig:causal_mask}
% \end{figure}
\begin{figure}[!tb]
    \centering
    \includegraphics[width=0.9\linewidth]{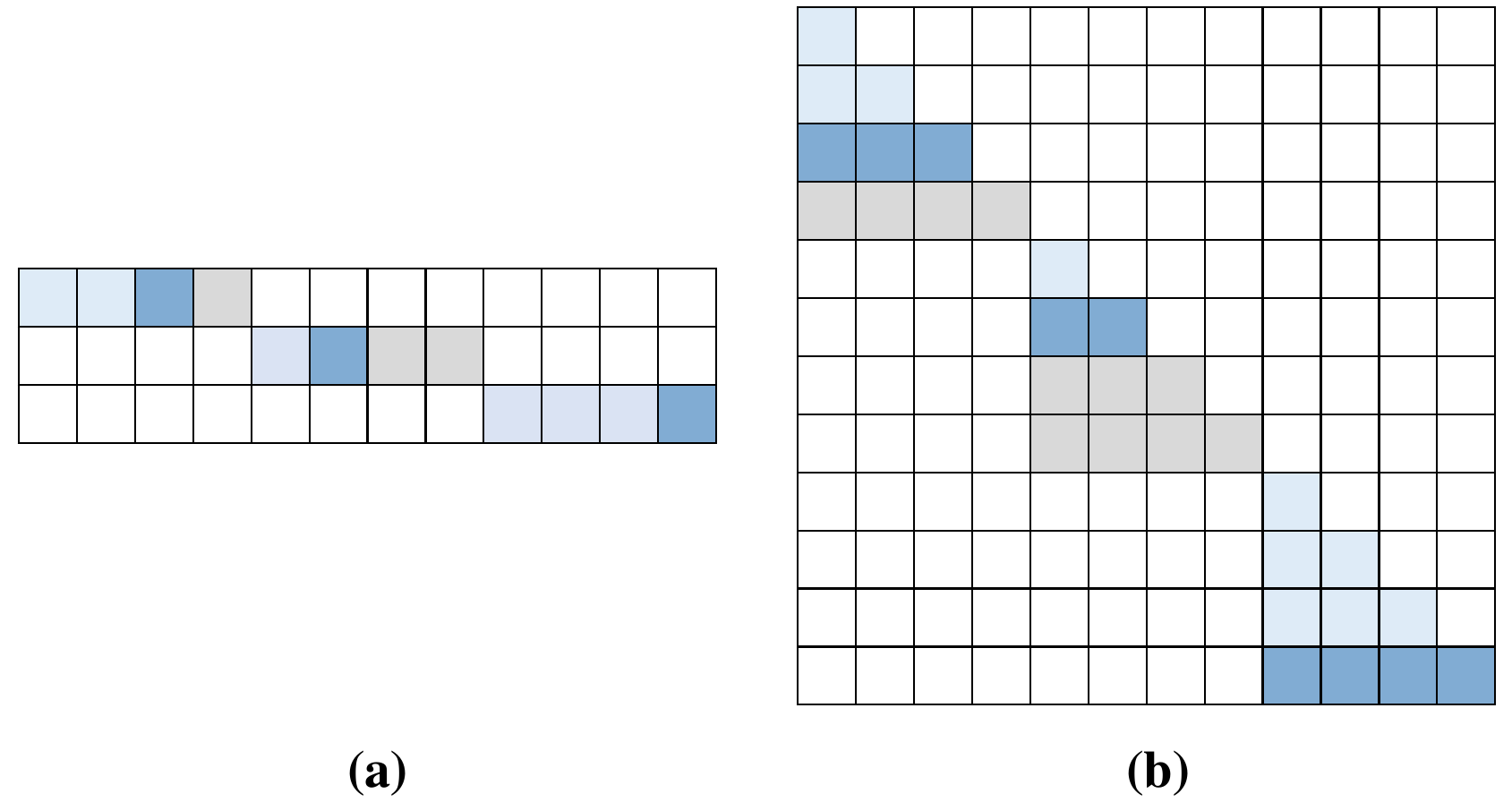}
    \caption{(a) Batch of suffix infilling temporal sequences; (b) The causal masking for the batch of suffix infilling temporal sequences. Cells in gray are padding index zero and are masked in the self-attention mechanism. } 
    \label{fig:causal_mask}
\end{figure}

Based on suffix infilling and temporal graph attention with self-loop operations, we further introduce the causal masking self-attention mechanism shown in Figure \ref{fig:attention_causal} to bridge the gap between the sequence modeling and the temporal message aggregation. The suffix infilling temporal sequence is formatted as a  sequence $\mathcal{X}_u(t)$. For the suffix infilling temporal sequences, we truncate the long sequences to the set length $l$ and pad the short sequences with the padding index zero:
\begin{equation}
    \small
    \mathcal{X}_v(t) = (\underbrace{v_{1}(t_1), \dots, v_{k}(t_k)}_k, v(t), \underbrace{0, \dots, 0}_{l-k-1}),
\end{equation}
where the maximum neighbor size $k$ is $l - 1$.  Figure \ref{fig:causal_mask}a shows a batch of temporal sequences with padding or truncation operation.

Based on the sequence $\mathcal{X}_v(t)$, we obtain the node hidden states of the sequence $\mathcal{X}_u(t)$ as
\begin{equation}
    \small
    \mathbf{E}_{node}(t)=[\mathbf{h}_{v_1}(t_1), \dots, \mathbf{h}_{v_k}(t_k), \mathbf{h}_{v}(t), \underbrace{\vec{0}, \dots, \vec{0}}_{l-k-1}]^{\top}\in\mathbb {R}^{l\times~d_v}
\end{equation}
by the node indices, where the $\mathbf{h}_{0}$ is the frozen embedding zeros, and so are the edge features~$\mathbf{E}_{edge}(t)$ and time embeddings~$\mathbf{E}_{time}(t)$. 

% \noindent\textbf{Edge embedding [optional].} The edge indices of the sequence $\mathcal{X}_u(t)$ are the indices of edges $\{(v(t), v_1(t_1)), \dots, (v(t), v_k(t_k))\}$, denoted as 
\begin{equation}
    \small
    \mathbf{E}_{edge}(t)=[\mathbf{e}_{1}(t_1), \dots, \mathbf{e}_{k}(t_k), \mathbf{e}_{self}(t),\underbrace{\vec{0}, \dots, \vec{0}}_{l-k-1}]^{\top}\in\mathbb {R}^{l\times~d_e},
\end{equation}
% where $\mathbf{e}_{self}(t)$ is a self-loop edge embedding and can be a vector of zeros, and $\vec{0}$ is the freezing embedding zeros.

% % There are two cases for the edge embedding: 
% % \begin{itemize}
% %     \item The edge embedding matrix has been learned by embedding methods (e.g. word2vec, text embedding models), and each embedding represents a real-world edge feature. Therefore, the edge embeddings are frozen and cannot be optimized.
% %     \item The edge embedding matrix is not learned, and each edge embedding
% %     is randomly initialized and optimized during the training process. 
% % \end{itemize}

% \noindent\textbf{Time embedding.} As proved in previous literatures~\cite{xu2019self,xu2020inductive}, we use a time encoder $\Phi$, which is a multi-layer perceptron (MLP) that encodes the time intervals information between the current node $v(t)$ and its sorted temporal neighbors $\overrightarrow{\mathcal{N}}_{v}(t)$ as 
\begin{equation}
    \small
    \mathbf{E}_{time}(t)=[\Phi(t-t_{1}), \dots, \Phi(t-t_{k}), \Phi(0),\underbrace{\vec{0}, \dots, \vec{0}}_{l-k-1}]^{\top} \in \mathbb{R}^{l\times~d_t},
\end{equation}
% where $\Phi(0)$ is the time embedding of the self-loop edge and can be a vector of zeros, and $\vec{0}$ is the freezing embedding zeros.

% For the node embedding, edge embedding, and time embedding introduced in Section~\ref{subsec:TGAT}, we can concatenate them as the input of the causal masking self-attention mechanism:
%  \in \mathbb{R}^{l\times~(d_v+d_e+d_t)}
We concatenate them as the input of the causal masking self-attention:
\begin{equation}
    \small
    \begin{aligned}
        \centering
        \mathbf{Z}_{v}(t) =  \mathbf{E}_{node}(t) \oplus \mathbf{E}_{edge}(t) \oplus \mathbf{E}_{time}(t),
    \end{aligned}
\end{equation}
% \begin{equation}
%     \begin{aligned}
%         \centering
%         \mathbf{Z}_{v}(t) = & \mathbf{E}_{node}(t) \oplus \mathbf{E}_{edge}(t) \oplus \mathbf{E}_{time}(t),\\ 
%         = &[\mathbf{h}_{v_1}(t_1)\oplus \mathbf{e}_{1}(t_1)\oplus \Phi(t-t_{1}), \dots, \mathbf{h}_{v_k}(t_k)\oplus \mathbf{e}_{k}(t_k)\oplus \\&\Phi(t-t_{k}),\mathbf{h}_{v}(t) \oplus \mathbf{e}_{self}(t) \oplus \vec{0},\underbrace{\vec{0},\dots, \vec{0}}_{l-k-1}]^{\top}, \\
%         = &[\mathbf{h}_{v_1}(t_1)\oplus \mathbf{e}_{1}(t_1)\oplus \Phi(t-t_{1}), \dots, \mathbf{h}_{v_k}(t_k)\oplus \mathbf{e}_{k}(t_k)\oplus \\&\Phi(t-t_{k}), \tilde{\mathbf{h}}_{v}(t),\underbrace{\vec{0},\dots, \vec{0}}_{l-k-1}]^{\top} \in \mathbb{R}^{l\times~(d_v+d_e+d_t)}, \\
%     \end{aligned}
% \end{equation}
% where $\tilde{\mathbf{h}}_{v}(t)$ denotes the vector $\mathbf{h}_{v}(t)$ padded with $\vec{0}\in \mathbb{R}^{d_e+d_t}$, 
where $\oplus$ also can denote the sum operation when the dimensions of the embeddings are the same and the $\mathbf{e}_{self}(t)$ is a frozen zero vector $\vec{0}$.
The last $l-k-1$ rows of $\mathbf{Z}_{v}(t)$ are padding zeros, which are all frozen in the the self-attention mechanism. 
The features, such as the number of neighbors and time quantification, can be easily integrated into the causal masking self-attention mechanism.
The query, key, and value of the transfomer decoder~\cite{vaswani2017attention} are calculated based on the hidden states of the sequence $\mathcal{X}_u(t)$ as follows:
\begin{equation}
    \centering
    \small
    \label{eq:qkv_causal}
    \begin{aligned}
        \centering
        \mathbf{Q}(t) = [\mathbf{Z}_{v}(t)]_{0:l}\mathbf{W}_{Q}, 
        \mathbf{K}(t)  = [\mathbf{Z}_{v}(t)]_{0:l}\mathbf{W}_{K},
        \mathbf{V}(t) =[\mathbf{Z}_{v}(t)]_{0:l}\mathbf{W}_{V}.
    \end{aligned}
\end{equation}
% \begin{equation}
%     \centering
%     \label{eq:qkv_causal}
%     \begin{aligned}
%         \centering
%         \mathbf{Q}(t) & = [\mathbf{Z}_{v}(t)]_{0:l}\mathbf{W}_{Q}, \\ 
%         \mathbf{K}(t) & = [\mathbf{Z}_{v}(t)]_{0:l}\mathbf{W}_{K},\\
%         \mathbf{V}(t)& =[\mathbf{Z}_{v}(t)]_{0:l}\mathbf{W}_{V}.
%     \end{aligned}
% \end{equation}
% to preserve the auto-regressive property.
The  masking matrix $\mathbf{M}\in \{0,1\}^{l \times~l}$ in the causal masking self-attention mechanism enables \textit{each element to attend to itself and all preceding elements}:

% aims to prevent the attention mechanism from attending to the future elements, and its formula is as:
\begin{equation}
    \small
    \mathbf{M}_{i} = [\underbrace{0,\dots,0}_{i}, \underbrace{-\infty , \dots, -\infty }_{l-i}], i\in \{1,\dots, l\},
\end{equation}
where $-\infty$ denotes the attention coefficient that is masked out, 0 keeps.

The self-attention mechanism with causal masking can be formulated as follows:
\begin{equation}
    \small
    \label{eq:causal_attention}
    \text{Attn}(\mathcal{X}_{v}(t)) = \text{softmax}(\frac{\mathbf{Q}(t)\mathbf{K}(t)^{\top}}{\sqrt{d}} + \mathbf{M})\mathbf{V}(t),
\end{equation}
where $d = d_v+d_e+d_t$. Figure~\ref{fig:causal_mask}b shows the causal masking self-attention for the batch of suffix infilling temporal sequences. 

Let $\mathbf{Z}_{v}^{'}(t)$ denote the output of a \textit{multi-layer} model obtained by stacking the \textit{multi-head} causal masking self-attention. The updated hidden states of the node $v$ at time $t$ can be selected by the row index $k$ as follows:
\begin{equation}
    \small
    \mathbf{h}_{v}^{'}(t) = [\mathbf{Z}_{v}^{'}(t)]_{k}.
\end{equation}
Then, $\mathbf{h}_{v}^{'}(t)$ can be used for the downstream tasks, such as the dynamic link prediction task~\cite{kumar2019predicting,rossi2020temporal,xu2020inductive, huang2024benchtemp}. 

% \textit{neighborhood representations}~\cite{xu2020inductive}

It can be derived that the classical self-attention mechanism of TGAT~\cite{xu2020inductive} introduced in Section \ref{subsec:TGAT} and the temporal graph attention variant with self-loop in Section \ref{eq:qkv_selfloop} are variants of the self-attention mechanism with a specific masking matrix. Therefore, the causal masking self-attention in the TGNN framework can be extended to various temporal graph attention mechanisms and static graph attention mechanisms that do not require time encoders. This approach establishes a unified training paradigm, integrating sequence modeling with temporal message aggregation in TGNNs.

% a unified training paradigm that bridges sequence modeling and temporal message aggregation of TGNNs.
% , and unifid the traning paradigm between the sequence modeling and the temporal message aggregation of TGNNs.

Furthermore, the causal masking self-attention mechanism can be implemented on top of the  flash-attention~\cite{dao2022flashattention, dao2023flashattention} and memory-efficient attention~\cite{rabe2021self}, to accelerate the training process of the TGNN framework.
Flash-attention utilizes tiling to minimize the memory reads and writes between GPU HBM and SRAM. Memory-efficient attention utilizes bucketing by chunking the query, key, and value matrices to reduce the memory consumption. 
Based on these two efficient acceleration mechanisms of causal masking self-attention, TF-TGN can be implemented effectively to speed up the training process, enabling the construction of efficient TGNNs using the Transformer decoder for large-scale dynamic graphs.

\section{Efficient Training}
In this section, we introduce how to efficiently train TF-TGN based on the Transformer decoder. We first introduce the parallel sampling strategy to efficiently convert the temporal graph to the T-CSR format and sample the temporal neighbors in parallel. Then, we describe the batch training and the distributed training strategies of TF-TGN to accelerate the training process.

\subsection{Parallel Sampling Strategy}
\label{subsec:sampling}
\begin{figure}[!tb]
    \centering
        \includegraphics[width=0.6\linewidth]{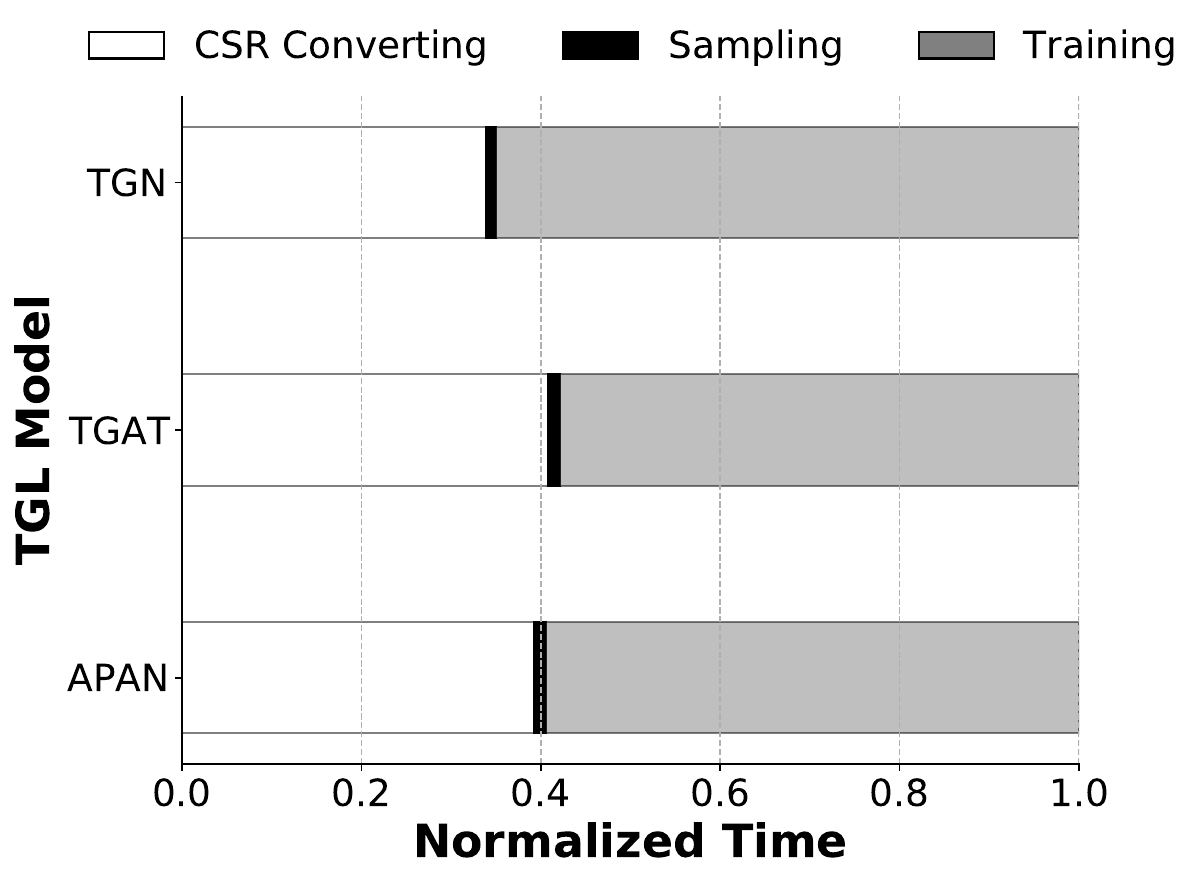}
        \caption{Comparison of the normalized time for the CSR converting, temporal neighbor sampling, and model training when training different TGNNs integrated with the TGL framework on the GDELT datasets.}
        \label{fig:normalized_time}
\end{figure}
\begin{algorithm}[!tb]
    \small
    % \footnotesize
    \caption{Parallel Sampling Strategy}
    \label{alg:sampling}
    % \KwData{The temporal interactions of a dynamic graph.}
    \KwIn{
    \\1. The temporal interactions $(u_{i}, v_{i}, t_{i})$ of a dynamic graph $G$, the indptr array $\mathbf{p}$, the edge idx array $\mathbf{e}$, the indices array $\mathbf{v}$, the time array $\mathbf{t}$; \\2. Batch input nodes $\mathbf{u_b}$ with batch timestamps $\mathbf{t_b}$.
    }
    \KwOut{
    % \begin{enumerate}[label=\arabic*.]
    %     \item The T-CSR representation of the dynamic graph;
    %     \item The sampled temporal neighbors of each node.
    % \end{enumerate}
    \\1. The T-CSR representation of the dynamic graph: $\mathbf{p}, \mathbf{e}, \mathbf{v}, \mathbf{t}$; \\2. The $k$ temporal neighbors of each node $u$ at time $t$,  $\mathcal{N}_u(t)$.
    }
    \BlankLine
    % \tcp{Count the number of edges for each source node}
    Initialize $\mathbf{p}$, $\mathbf{e}$, $\mathbf{v}$, $\mathbf{t}$;\\
    % \#pragma omp parallel for\\
    \For{$i \leftarrow 1$ \KwTo $n$ in parallel}{
        % \#pragma omp atomic\\
        $\mathbf{p}[\mathbf{v}[i] + 1]$++\tcp*{atomic access}
        \tcp{dst -> src}
        % \#pragma omp atomic\\
        % $\mathbf{p}[\mathbf{u}[i] + 1]$++\tcp*{dst -> src [optional]}
    }
    % $\mathbf{p}[0] = 0$ \tcp*{the first element is 0}
    
    \For{$i \leftarrow 1$ \KwTo $|\mathbf{p}|$}{
        $\mathbf{p}[i]~+= \mathbf{p}[i-1]$ \tcp*{cumulative sum for indptr}
    }
    $\mathbf{p}'$ = copy$(\mathbf{p})$ \tcp*{copy the indptr array}
    % \tcp{Fill edge idx, time, and indices}
    % \#pragma omp parallel for\\
    \For{$i \leftarrow 1$ \KwTo $n$ in parallel}{ 
        $e$ = $\mathbf{e}[i]$;~$u$ = $\mathbf{u}[i]$; ~$v$ = $\mathbf{v}[i]$;$t$ = $\mathbf{t}[i]$;\\

        % \#pragma omp atomic capture\\
        $\mu$ = $\mathbf{p}'[u]$++ \tcp*{atomic access}
        $\mathbf{e}[\mu]$ = $e$;~$\mathbf{v}[\mu]$ = $v$;~$\mathbf{t}[\mu]$ = $t$;\\
        
        \tcp{dst -> src }
        % \#pragma omp atomic capture\\
        % $j_v$ = $\mathbf{p}'[u]$++;\\
        % $\mathbf{e}[j_v]$ = $e$;~$\mathbf{v}[j_v]$ = $v$;~$\mathbf{t}[j_v]$ = $t$;\\

    }
    % \tcp{Rearrange indptr according to timestamps}
    % \#pragma omp parallel for\\
    \For{$i \leftarrow 0$ \KwTo $|\mathbf{p}|-1$ in parallel}{
        $m = \mathbf{p}[i]$; $n = \mathbf{p}[i+1]$;\\
        % \If{$n-m<2$}{continue}
        
        $\vec{l}$ = sort($\mathbf{t}[m:n]$) \tcp*{sorting and copying}
        $\mathbf{e}[m:n]$ =  $\mathbf{e}[m:n][\vec{l}]$; $\mathbf{v}[m:n]$ =  $\mathbf{v}[m:n][\vec{l}]$; $\mathbf{t}[m:n]$ =  $\mathbf{t}[m:n][\vec{l}]$;\\
        % $\mathbf{e}[m:n]$ =  $\mathbf{e}[m:n][\vec{j}]$;\\
        % $\mathbf{v}[m:n]$ =  $\mathbf{v}[m:n][\vec{j}]$;\\
        % $\mathbf{t}[m:n]$ =  $\mathbf{t}[m:n][\vec{j}]$;\\
    }
    % \tcp{Sample $k$ temporal neighbors}
    % \#pragma omp parallel for \tcp*{sampling in parallel}
    % \#pragma omp parallel for\\
    \For{$i \leftarrow 0$ \KwTo $|\mathbf{u_b}|$ in parallel}{
        $u$ = $\mathbf{u_b}[i]$;~$t$ = $\mathbf{t_b}[i]$;\\
        
        % \tcp{Binary search the index by the timestamp $\mathbf{t_b}[i]$}
        $m  $ = binarysearch$(\mathbf{t}[\mathbf{p}[u]:\mathbf{p}[u+1]],  t)$ \\
        $\mathcal{N}_{u}(t)$ = sample$(\mathbf{v}[\mathbf{p}[u]:\mathbf{p}[u+1]][0:m], k)$
    }
    % \Return{$\mathbf{p}, \mathbf{e}, \mathbf{v}, \mathbf{t}, \mathcal{N}_{u}(t)$};
\end{algorithm}
Pioneering TGNNs store temporal neighbors using dictionaries or lists, which are inefficient and memory-intensive. The compressed sparse row (CSR) representation  is an efficient data structure for temporal neighbor sampling in dynamic graphs~\cite{zhou2022tgl}. However, converting the temporal interactions chronologically of a large-scale dynamic graph into temporal CSR (T-CSR) format is time-consuming. As shown in Figure \ref{fig:normalized_time}, CSR format conversion can constitute more than 30\% of the training time,
which limits the scalability of the training process on large-scale temporal graphs.

We propose a parallel sampling strategy detailed in Algorithm \ref{alg:sampling}, which is implemented in C++ with OpenMP integration~\cite{chandra2001parallel}. This approach leverages \textit{concurrent threads} and \textit{atomic access} mechanisms to enhance the efficiency of converting CSR structures and performing temporal neighbor sampling. 
The algorithm proceeds as follows: First, we count the number of edges for each node in parallel using atomic access. Next, we compute the cumulative sum for indptr. Then, we populate the edge indices, timestamps, and node indices in parallel with atomic access. After that, we sort the edge indices, timestamps, and node indices according to the timestamps in parallel. Finally, we sample $k$ temporal neighbors for each node in parallel.
Refer to Appendix \ref{apd:sampling} for complexity analysis.

% based on \textit{concurrent threads} and \textit{access atomically} to accelerate the CSR converting and temporal neighbor sampling. 

\subsection{Training}
\label{subsec:batch_traning}
\begin{figure}[!tb]
    \centering
    \includegraphics[width=0.7\linewidth]{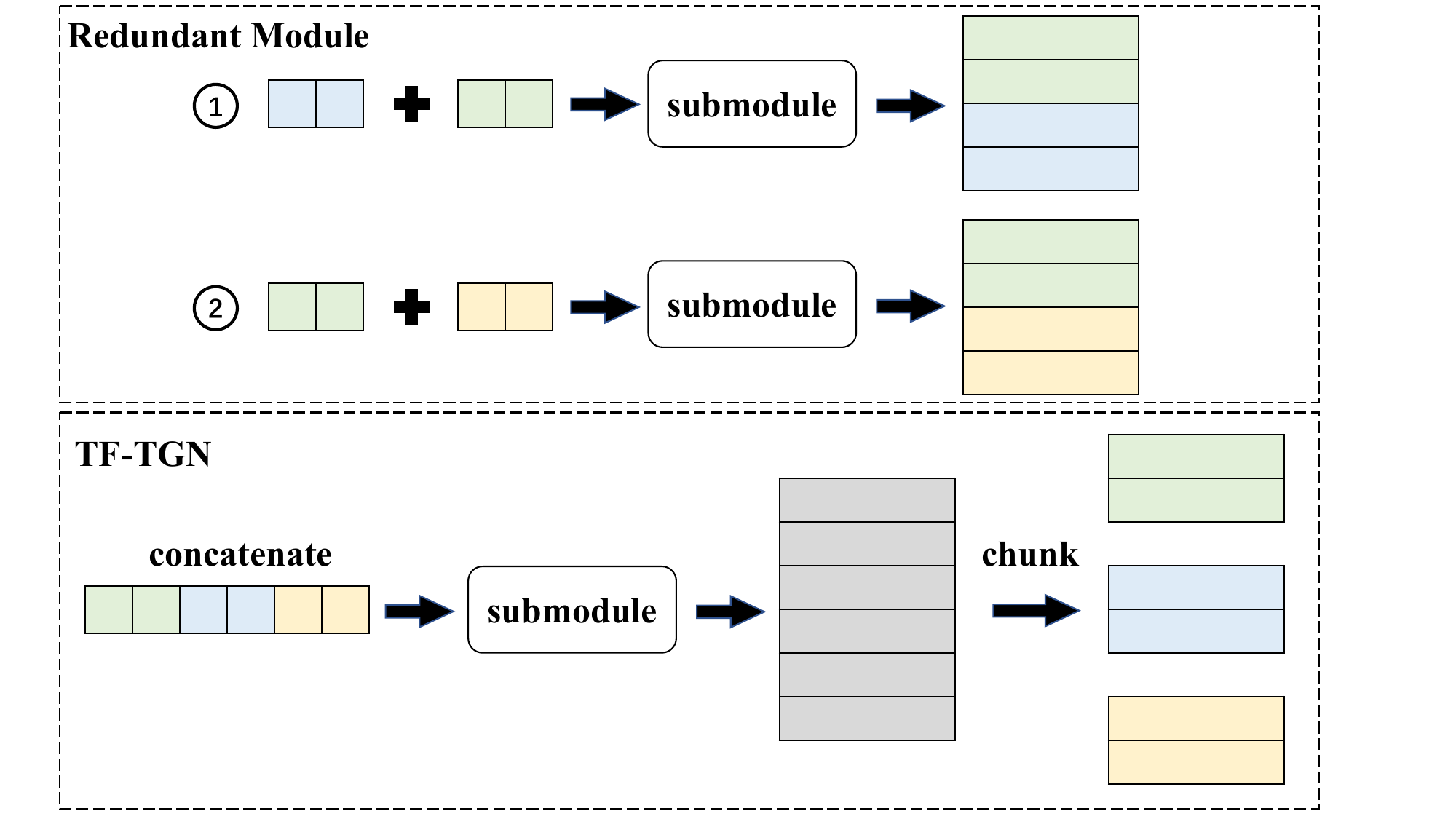}
    \caption{The batch training strategy of TGNNs.}
    \label{fig:batch}
\end{figure}

\noindent\textbf{Batch Training.} Using a submodule repeatedly for calculations within a single training forward pass is inefficient.
Furthermore, model parameters and data are distributed across multiple GPU devices in a distributed training setup. Replicating the same submodule with different inputs within a single forward pass can lead to errors and is thus not permitted~\cite{paszke2019pytorch, rasley2020deepspeed, FairScale2021}. This redundant module is prevalent and neglected in previous studies, which mainly focus on single-node training. We employ a batch training strategy to address this issue through concatenation and chunking operations as shown in Figure \ref{fig:batch}. 
We process the model input by concatenating sequences of source nodes, target nodes, and negative samples into a single long sequence within a batch. This sequence is processed in a single forward pass, after which corresponding features are obtained through a chunking operation. In this strategy, each submodule executes once per input during the forward pass, accumulating gradients across multiple inputs and  expediting the training process.

% This method facilitates seamless integration into distributed training frameworks.

\noindent\textbf{Distributed Training}
\label{subsec:distributed}
We train the TF-TGN in a distributed manner to significantly accelerate the training process. 
% However, training a dynamic graph composed of a large number of temporal interactions in a distributed manner  suffers from the temporal dependency of node states in intra-batches and inter-batches~\cite{kumar2019predicting,zhou2022tgl}. 
For a sequence of $n$ temporal interactions $S = \{s_{1}, \dots, s_{n}\}$, and  $m$ GPU nodes, the distributed data loader distributes a $b$ mini-batch size of temporal interactions that are consecutive on the edge index to each GPU node at a time ${B}_{\mu} = \{s_{ib}, \dots, s_{(i +1)b-1}\}
$, $i\in \mathbb{Z}^{+}$, while $m\times b$ temporal interactions  for all $m$ GPUs ${B} = \{{B}_{\mu}^{1}, \dots, {B}_{\mu}^{m}\}$. 
% In the distributed data parallel training of TGNNs, the intra-batch dependencies are the dependencies between the temporal interactions in the same mini-batch, i.e., dependencies in ${B}_{\mu}$,  and the inter-batch dependencies are the dependencies between the temporal interactions in different mini-batches but in the same batch, i.e., dependencies in ${B}$. 
The parameters $\theta$ of the TGNN model are updated by the stochastic gradient descent (SGD) algorithm in 
a distributed manner with a learning rate $\eta$ on $m$ GPU nodes, 

\begin{equation}
    \label{eq:gradient_descent}
    \theta^{(t+1)} \leftarrow \theta^{(t)}-\eta (\frac{1}{m} \sum_{i=1}^m \nabla \theta_i^{(t)}),
\end{equation}
where $\nabla \theta_i^{(t)})$ is the gradient  of the loss with respect to the parameters $\theta$ on the $i$-th GPU node at step $t$.

\section{Experiments}
\subsection{Experiment Settings}
\textbf{Environment.} All experiments are conducted on a server with 8 NVIDIA A800 80GB PCIe GPUs, 1TB memory, and 2 Intel(R) Xeon(R) Platinum 8380 CPUs @ 2.30GHz with 80 cores and 160 threads. Each two GPUs is interconnected by NVIDIA NVLink 400GB/s. The P2P PCIe bandwidth is 37GB/s. The server is running on Ubuntu 20.04.6 LTS with CUDA 12.2. The parallel sampling strategy in Algorithm \ref{alg:sampling} is implemented in C++11 and OpenMP and can be integrated with PyBind11~\cite{pybind11, chandra2001parallel}.
TF-TGN is implemented using PyTorch FSDP and xFormers, which incorporate memory-efficient attention and flash-attention~\cite{xFormers2022}.

% Table generated by Excel2LaTeX from sheet 'LP任务'
\begin{table}[!tb]
    \small
    \centering
    \caption{Statistics of the datasets.}
    \setlength{\tabcolsep}{4pt}
      \begin{tabular}{lrrrrr}
      \toprule
      Dataset & $|V|$ & $|E|$ &  $t_{max}$ & $d_{v}$ & $d_{e}$ \\
      \midrule
      UCI   & 1,899  & 59,835   & 16,736,181  & 172   & 100  \\
      Wikipedia  & 9,227  & 157,474     & 2,678,373  & 172   & 172  \\
      Reddit & 10,984  & 672,447     & 2,678,390  & 172   & 172  \\
      MOOC  & 7,144  & 411,749     & 2,572,086  & 172   & 4  \\
      LastFM & 1,980  & 1,293,103      & 137,107,267  & 172   & 2  \\
      Wiki-Talk  & 1,140,149  & 7,833,139      & 200,483,882  & 172   & 172  \\
      Stack-O.  & 2,601,977  & 63,497,049     & 239,699,627  & 172   & 172  \\
      GDELT & 16,682  & 191,290,882     & 175,283  & 413   & 186  \\
      MAG   & 72,508,661  & 1,297,748,926      & 120   & 768   & 172  \\
      \bottomrule
      \end{tabular}%
    \label{tab:datasets}%
  \end{table}%

\noindent\textbf{Datasets.} Table \ref{tab:datasets} presents statistics of the selected experimental datasets from diverse domains with up to billions of edges. We convert the temporal interactions of graphs into the T-CSR representation for neighbor sampling and the TGNN input format and partition the temporal interactions chronologically into three sets based on timestamps: the training set (70\%), the validation set (15\%), and the test set (15\%) adhering to the standardized procedure outlined in ~\cite{kumar2019predicting,rossi2020temporal,zhou2022tgl, gao2024etc}. Refer to Appendix \ref{apd:datasets} for details.

\begin{table*}[!tb]
    \centering
    \small
    \caption{Comparison of ROC AUC results on the dynamic link prediction task. The TGAT, TGN, and APAN are integrated with the TGL and ETC training frameworks. The best results in each column are colored in \textcolor{blue}{bold blue}. We highlight the top three results in each row using varying shades of blue.}
        \begin{tabular}{clrrrrrrrrr}
        \toprule
        \multirow{2}{*}{Framework} & \multirow{2}{*}{Model} & \multicolumn{9}{c}{Dataset} \\
    \cmidrule{3-11}          &       & UCI   & Wikipedia  & Reddit & MOOC  & LastFM & Wiki-Talk & Stack-O. & GDELT & MAG \\
        \midrule
        \multirow{3}[2]{*}{TGL} & \multicolumn{1}{l}{TGN} & \second0.8264  & \third0.9846  & \first{0.9965 } & \first{0.9959 } & \second0.8494  & \first{0.9654 } & \first{0.8991 } & \second0.9614  & \second0.8229  \\
                & \multicolumn{1}{l}{TGAT} & 0.7683  & 0.9508  & 0.9838  & 0.9865  & 0.7348  & 0.9041  & 0.8037  & 0.9491  & 0.6250  \\
                & \multicolumn{1}{l}{APAN} & 0.6900  & \second0.9864  & 0.9828  & \third0.9952  & 0.5729  & \second0.9601  & \second0.8859  & 0.9513  & 0.8076  \\
          \midrule
          \multirow{3}[2]{*}{ETC} & \multicolumn{1}{l}{TGN} & 0.5888  & 0.9689  & \second0.9940  & \second0.9953  & \third0.8362  & \third0.9501  & 0.6974  & \third0.9613  & \third0.8211  \\
                & \multicolumn{1}{l}{TGAT} & \third0.7816  & 0.8533  & 0.9575  & 0.9694  & 0.7197  & 0.8856  & 0.5956  & 0.9392  & 0.6086  \\
                & \multicolumn{1}{l}{APAN} & 0.6634  & 0.9245  & 0.9130  & 0.9626  & 0.7260  & 0.9146  & 0.6698  & 0.9377  & 0.8074  \\
          \midrule
          &TF-TGN & \first{0.8762 } & \first{0.9898 } & \third0.9854  & 0.9933  & \first{0.8838 } & 0.9497  & \third0.8415  & \first{0.9618 } & \first{0.8366 } \\
        \bottomrule
        \end{tabular}%
    \label{tab:AUC}%
    \end{table*}%
    \begin{figure*}[!t] % Use figure* to span both columns
        \begin{minipage}[!t]{0.48\textwidth} % Adjust the width as needed
            \captionsetup{type=table}
            \centering
            \small
            \setlength{\tabcolsep}{3.5pt}
            \caption{The per-epoch training time (\textit{seconds}) of TGNN models on relatively small datasets. The best results in the table are highlighted in \textcolor{blue}{bold blue}. The second-best results are highlighted in \textcolor{lightblue}{light blue}. The speedup is calculated as the ratio of the best result to the second-best result.}
            % The speedup is calculated as the ratio of the training time of the classical TGAT model to that of TF-TGN.
              \begin{tabular}{clrrrrr}
                \toprule
                \multirow{2}{*}{Framework} & \multirow{2}{*}{Model} & \multicolumn{5}{c}{Dataset} \\
          \cmidrule{3-7}          &       & UCI   & Wikipedia  & Reddit & MOOC  & LastFM \\
          \midrule
        %   &TGAT & 57.49 & 109.24 & 515.64 & 256.26 & 882.36 \\
            %   \midrule
              \multirow{3}[0]{*}{TGL} & \multicolumn{1}{l}{TGN} & 3.18  & 3.20  &\second 8.30  & \second5.96  & \second15.92  \\
                    & \multicolumn{1}{l}{TGAT} & 3.05  & 6.19  & 22.63  & 16.30  & 46.57  \\
                    & \multicolumn{1}{l}{APAN} & \second 1.66  & 3.65  & 13.28  & 4.61  & 24.87  \\
                    \midrule
              \multirow{3}[0]{*}{ETC} & \multicolumn{1}{l}{TGN} & 1.33  & 3.52  & 13.14  & 9.44  & 20.99  \\
                    & \multicolumn{1}{l}{TGAT} & 2.31  & \second2.71  & 10.68  & 7.18  & 17.23  \\
                    & \multicolumn{1}{l}{APAN} & 1.90  & 3.16  & 10.40  & 6.09  & 16.86  \\
                    \midrule
                    % &TF-TGN & \textbf{0.99} (58.02$\times$) & \textbf{1.24} (87.75$\times$) & \textbf{5.05} (102.17$\times$) & \textbf{3.43} (74.69$\times$) & \textbf{10.31} (85.55$\times$) \\
                    &TF-TGN & \first{0.25}  & \first{1.24}  & \first{5.05}  & \first{3.43}  & \first{10.31}  \\
                    % &Speedup & 58.02$\times$  & 87.75$\times$  & 102.17$\times$  & 74.69$\times$  & 85.55$\times$  \\
                    &Speedup & 6.56$\times$  & 2.19$\times$  & 1.64$\times$  & 1.74$\times$  & 1.54$\times$  \\
                    % 1.676767677	2.185483871	1.643564356	1.737609329	1.544131911
            \bottomrule
              \end{tabular}%
            \label{tab:train_time_small}%
        \end{minipage}
        \hfill
        \begin{minipage}[!t]{0.48\textwidth} % Adjust the width as needed
            \captionsetup{type=table}
            \centering
            \small
            \setlength{\tabcolsep}{4pt}
            \caption{The training time (\textit{seconds}) per million edges for TGNN models on large-scale datasets. The best results in the table are highlighted in \textcolor{blue}{bold blue}. The second-best results are highlighted in \textcolor{lightblue}{light blue}. The speedup is calculated as the ratio of the best result to the second-best result.}
              \begin{tabular}{clrrrr}
              \toprule
              \multirow{2}{*}{Framework} & \multirow{2}{*}{Model} & \multicolumn{4}{c}{Dataset} \\
          \cmidrule{3-6}          &       & Wiki-Talk & Stack-O. & GDELT & MAG \\
              \midrule
              \multirow{3}[2]{*}{TGL} & \multicolumn{1}{l}{TGN} & 10.42  & 7.75  & 9.52  & 16.98  \\
                    & \multicolumn{1}{l}{TGAT} & \second9.50  & \second5.64  & \second7.03  & 14.93  \\
                    & \multicolumn{1}{l}{APAN} & 10.39  & 10.56  & 7.52  & \second14.78  \\
              \midrule
              \multirow{3}[2]{*}{ETC} & \multicolumn{1}{l}{TGN} & 30.05  & 29.91  & 15.04  & 23.02  \\
                    & \multicolumn{1}{l}{TGAT} & 34.55  & 30.67  & 26.33  & 26.07  \\
                    & \multicolumn{1}{l}{APAN} & 16.99  & 19.94  & 11.53  & 16.38  \\
              \midrule
              &TF-TGN & \first{3.67 } & \first{3.99 } & \first{4.54 } & \first{4.58 } \\
              &Speedup & 2.59$\times$  & 1.41$\times$  & 1.55$\times$  & 3.26$\times$  \\
            %   2.588555858	1.413533835	1.54845815	3.259825328
              \bottomrule
              \end{tabular}%
            \label{tab:train_time_large}%
        \end{minipage}
      \end{figure*}

\noindent{\textbf{Baselines}.} We compare the proposed TF-TGN with the state-of-the-art methods: TGN, TGAT, and APAN integrated with the efficient training frameworks TGL and ETC~\cite{zhou2022tgl, gao2024etc}. 
We utilize open-source implementations of baseline models and optimize hyperparameters to achieve optimal performance. The hybrid CPU-GPU data loading framework is employed for TGNNs training on large-scale dynamic graphs~\cite{zhou2022tgl, gao2024etc}. The neighbor sampling strategy of ETC is derived from TGL. Therefore, we compare the efficiency of Algorithm \ref{alg:sampling}'s neighbor sampling approach with the state-of-the-art sampling framework introduced in TGL~\cite{zhou2022tgl}. We focus on the  link prediction task. ROC AUC is selected as the evaluation metric.  Refer to Appendix \ref{apd:baselines} for details.

\noindent\textbf{Hyperparameters.} For the neighbor sampling, we set the maximum neighbor size $k$ from the set $\{2 \sim 12, 16, 32, 64, 128\}$ and adopt the recent sampling strategy, with the sampling layer maintained as 1, consistent with previous studies~\cite{zhou2022tgl, gao2024etc}. The length $l$ of the temporal sequence associated with neighbors is $|\mathcal{X}_v(t)|$. The hidden dimensions of node features $d_v$ and edge features $d_e$ are shown in Table \ref{tab:datasets}. We select the dimension of time embedding $d_t$ among \{64, 100, 128, 256\}. The number of heads in the multi-head attention mechanism ranges from \{2,4,8,16\}, with an embedding size per head chosen from \{128,256\}. We stack the multi-head causal masking Attention layer using values from $\{1\sim~8\}$.
The batch size is chosen from \{256, 512, 600, 1500, 2000\}. We use the Adam optimizer with a learning rate of 0.0001, and the dropout rate is selected from \{0.1, 0.2\}. In distributed training, the maximum number of GPUs is set to 8.
Mixed precision BF16~\cite{micikevicius2017mixed} is used to accelerate the training process. ZeRO-2, i.e., the strategy of sharding gradients and optimizer states during computation, is utilized to reduce memory consumption and enhance training efficiency~\cite{FairScale2021}. CPU offloading is employed to transfer parameters to the CPU when they are not involved in GPU computations.

\subsection{Main Results}

\subsubsection{Performance}
We conduct extensive experiments to evaluate the performance of the proposed TF-TGN on the  dynamic link prediction task. We compare the TF-TGN with the state-of-the-art methods TGAT, TGN, and APAN integrated with the TGL and ETC  frameworks to facilitate efficient training on large-scale dynamic graphs. The ROC AUC results are shown in Table \ref{tab:AUC}. Among the five relatively small datasets, TF-TGN achieves the best performance on UCI, Wikipedia, and LastFM datasets. On the Reddit dataset, TF-TGN achieves the third-best performance, only slightly behind the TGL-TGN and ETC-TGN models. On the MOOC dataset, TF-TGN achieves competitive performance, ranking fourth, nearly equivalent to the best performance.
On the large-scale datasets, TF-TGN achieves the best performance on the GDELT and MAG datasets and the third-best performance on the Stack-O. dataset. 
On the Wiki-Talk dataset, TF-TGN attains the fourth-best performance, slightly behind the third-ranked model, ETC-TGN.
The integration of TGN with TGL and ETC training frameworks obtains the best or second-best performance on 4 out of 9 datasets, highlighting the effectiveness of the memory-based TGNN approach. Additionally, APAN integrated with the TGL framework achieves top-three performance on 4 out of 9 datasets.
The results demonstrate that the proposed TF-TGN can effectively handle large-scale dynamic graphs with up to billions of edges and achieve state-of-the-art performance on the dynamic link prediction task.  
\begin{table}[!tbp]
    \small
        \centering
        \caption{Comparison of the time (\textit{seconds}) required to convert the temporal interactions of a dynamic graph into the T-CSR representation and the speedup achieved between the SOTA method of TGL and Algorithm \ref{alg:sampling} implemented by C++ and OpenMP on the datasets. Reverse means that the reverse edges are considered.}
        \setlength{\tabcolsep}{2pt}
        \begin{tabular}{lrrrr}
            \toprule
            \multirow{2}[4]{*}{Dataset} & \multicolumn{2}{c}{\textbf{No reverse}} & \multicolumn{2}{c}{\textbf{Reverse}} \\
            \cmidrule(lr){2-3}  \cmidrule(lr){4-5} 
            & TGL  & TF-TGN   & TGL & TF-TGN   \\
            \midrule
            UCI   & 3.06  & 0.03 (115.59$\times$) & 3.41  & 0.04 (80.2$\times$) \\
            Wikipedia  & 7.78  & 0.04 (182.31$\times$) & 9.26  & 0.05 (174.17$\times$) \\
            Reddit & 33.29  & 0.05 (630.24$\times$) & 40.03  & 0.23 (174.74$\times$) \\
            MOOC  & 20.94  & 0.05 (453.89$\times$) & 23.97  & 0.25 (96.3$\times$) \\
            LastFM & 50.32  & 0.15 (341.41$\times$) & 73.89  & 0.3 (246.85$\times$) \\
            Wiki-Talk  & 288.31  & 2.21 (130.59$\times$) & 316.04  & 2.63 (120.04$\times$) \\
            Stack-O. & 2,496.47  & 2.35 (1064.17$\times$) & 2,081.87  & 5.92 (351.38$\times$) \\
            GDELT & 7,084.81  & 47.79 (148.24$\times$) & 11,121.56  & 105.85 (105.07$\times$) \\
            MAG   & 39,325.73  & 26.82 (1466.45$\times$) & 43,258.30  & 62.93 (687.36$\times$) \\
            \midrule
            AVG. speedup     &   & 503.65$\times$  &  & 226.23$\times$  \\
            \bottomrule
            \end{tabular}%
        \label{tab:converting}%
\end{table}

\begin{figure}[!tbp]
    \centering
        \includegraphics[width=0.9\linewidth]{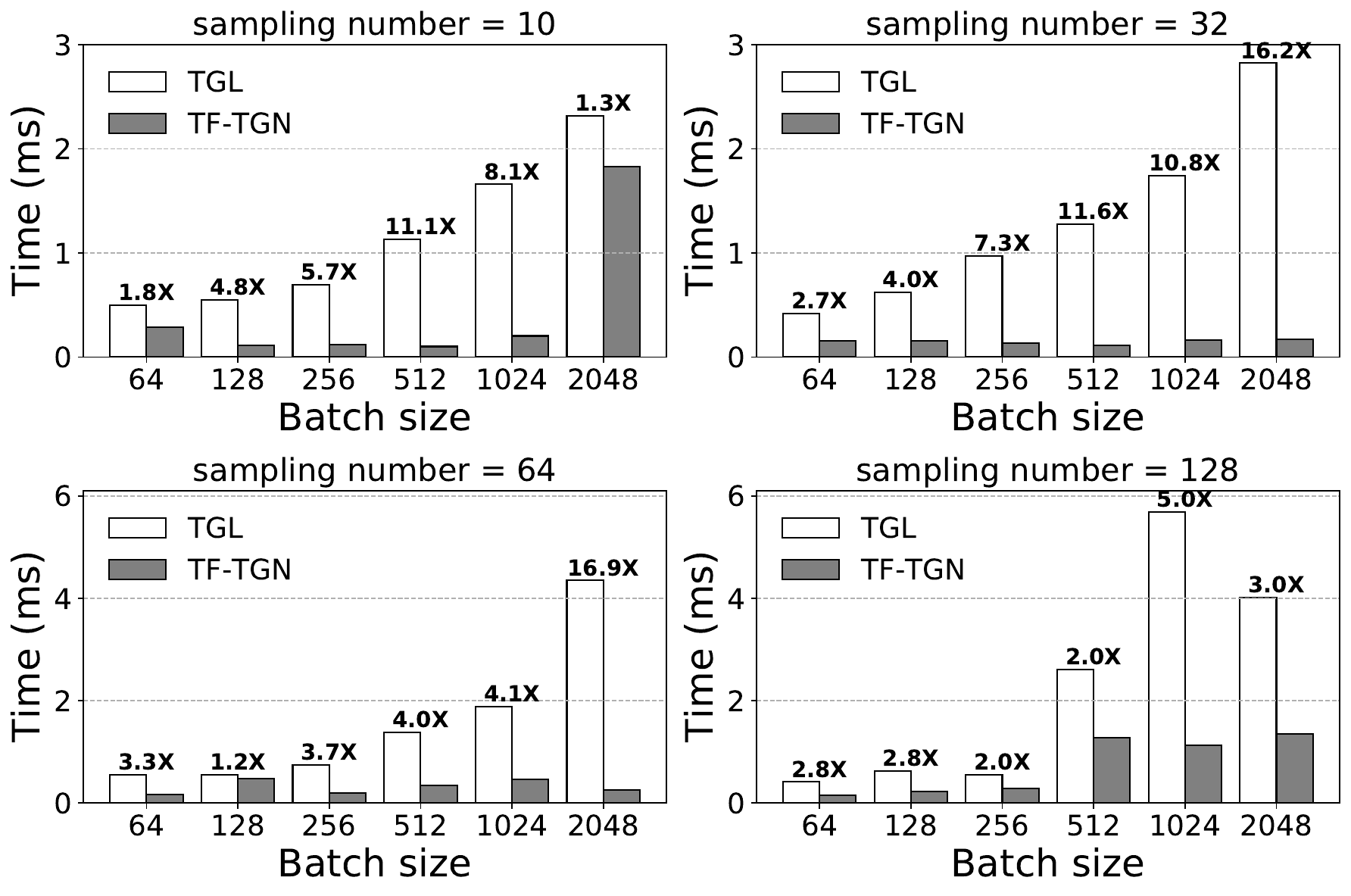}
        \caption{Comparison of sampling time and speedup between Algorithm~ 
        \ref{alg:sampling} and TGL on the Reddit dataset at different sampling numbers and batch sizes. The numbers above each group of bars represent the speedup achieved by Algorithm \ref{alg:sampling} compared to TGL.}
        \label{fig:sampling_time}
\end{figure}

\subsubsection{Efficiency} We evaluate the training time per epoch of the TGNN models on the relatively small datasets, as shown in Table \ref{tab:train_time_small}. The TF-TGN model achieves the best performance on all datasets compared to the SOTA TGNN modes integrated with the existing efficient training frameworks.
The average speedup of TF-TGN compared to the second-best results is 2.73$\times$. We further evaluate the training time per million edges of the TGNN models on the large-scale datasets, as shown in Table \ref{tab:train_time_large}. The TF-TGN model achieves the best performance on all datasets as well. The average speedup of TF-TGN compared to the second-best results is 2.20$\times$. Additionally, the TGN model integrated with the TGL training framework achieves the second-best results on 3 out of 5 relatively small datasets, demonstrating the effectiveness of the memory-based TGNN approach in handling small temporal graphs. In contrast, on the large-scale datasets, the TGAT model integrated with the TGL training framework achieves the second-best results on 3 out of 4 datasets, highlighting the effectiveness of the attention-based TGNN approach in handling large-scale temporal graphs.

\begin{figure}[!tb]
    \centering
    \includegraphics[width=0.9\linewidth]{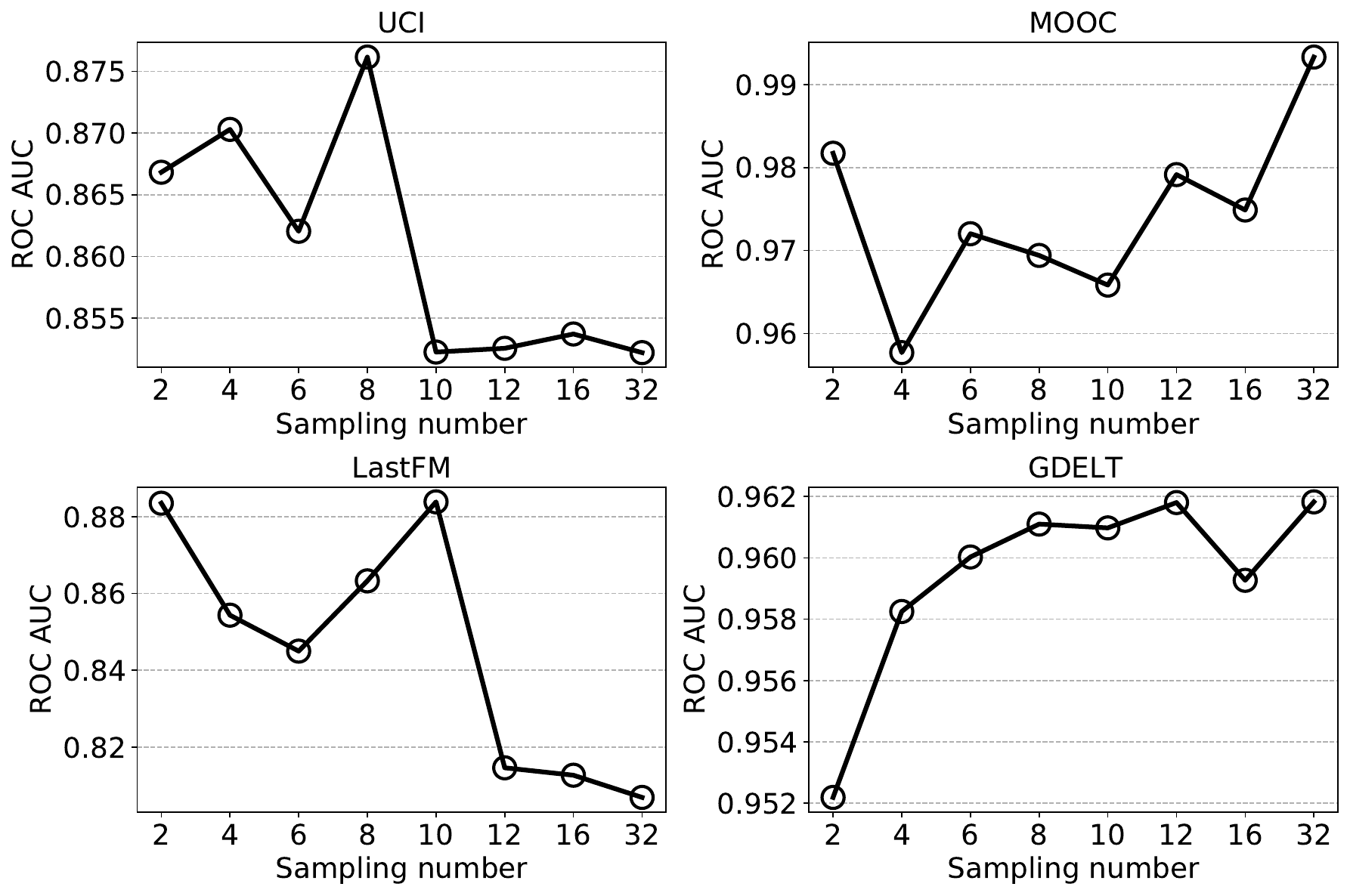}
    \caption{The ROC AUC results of TF-TGN on the dynamic link prediction task with different sampling numbers.}
    \label{fig:neighbor_num}
\end{figure}

\subsection{Sampling Efficiency}
We evaluate the efficiency of the parallel sampling strategy outlined in Algorithm \ref{alg:sampling} on the Intel Xeon(R) Platinum 8380 CPU @ 2.30GHz with up to 160 threads and 1TB memory. First, we compare the time required to convert the temporal interactions of a dynamic graph into the T-CSR representation and the speedup achieved between the SOTA method of TGL and Algorithm \ref{alg:sampling}. The results are shown in Table \ref{tab:converting}. We can observe that the classical approach of TGL performs well when the temporal graph is small, but the time required to convert the temporal interactions into the T-CSR representation increases significantly as the size of the temporal graph grows. For example, excluding reverse edges, TGL takes 3.06 seconds on the UCI dataset containing sixty thousand edges, whereas it requires 39,325.73 seconds on the MAG dataset, which includes billions of edges, highlighting challenges in scaling to large-scale dynamic graphs. In contrast, the time required by Algorithm \ref{alg:sampling} is only 0.03s on the UCI dataset, achieving a speedup of 115.59$\times$. The time required by Algorithm \ref{alg:sampling} is only 26.82s on the MAG dataset, achieving a speedup of 1466.45$\times$. The average speedup of Algorithm \ref{alg:sampling} compared to the existing SOTA method of TGL is 503.65$\times$ when reverse edges are not considered, and 226.23$\times$ when they are. The results demonstrate that Algorithm \ref{alg:sampling} can significantly accelerate the conversion of temporal interactions into the T-CSR representation and effectively handle large-scale dynamic graphs with up to billions of edges.

Furthermore, we conduct a comprehensive comparison of the sampling time and speedup between Algorithm \ref{alg:sampling} and TGL using the Reddit dataset across various sampling numbers and batch sizes, as illustrated in Figure \ref{fig:sampling_time}. The sampling number is selected among \{10, 32, 64, 128\}, and the batch size is chosen from \{64, 128, 256, 512, 1024, 2048\}. When the sampling number is 10, which is also the number used for training in TGL, Algorithm \ref{alg:sampling} shows significant speedup across all batch sizes, achieving the highest speedup of 11.1$\times$ with a batch size of 512. The average speedup at a sampling number of 10 is 5.46$\times$. When the sampling numbers are 32, 64, and 128, the situation mirrors that of sampling number 10, with average speedups of 8.75$\times$, 5.51$\times$, and 2.94$\times$, respectively.

\subsection{Additional Experiments}

\subsubsection{Performance with Sampling Number}

\begin{figure}[!tb]
    \centering
    \includegraphics[width=0.9\linewidth]{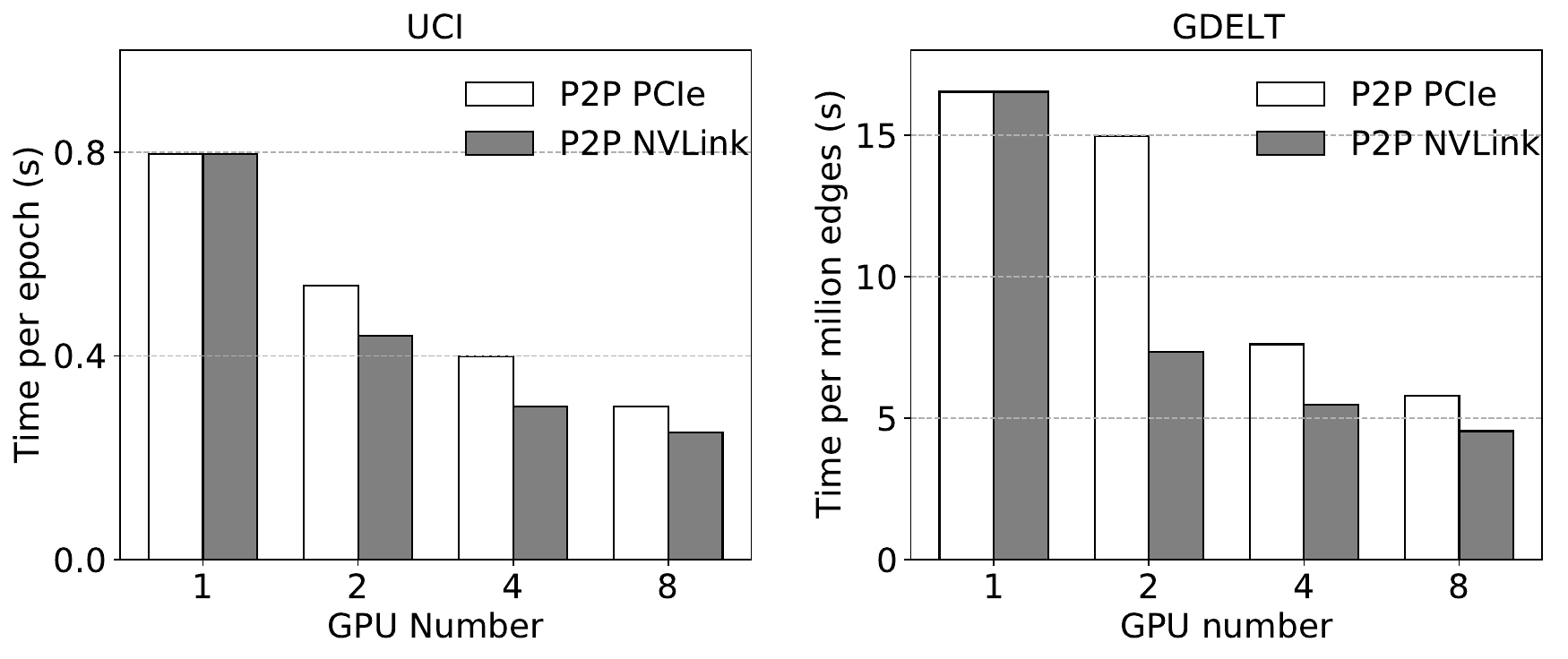}
    \caption{The training time with different numbers of GPUs on the UCI and GDELT datasets.}
    \label{fig:time_gpu}
\end{figure}
We evaluate the impact of the sampling number on the performance of TF-TGN on the dynamic link prediction task. 
The ROC AUC results of TF-TGN with different sampling numbers across the UCI, MOOC, LastFM, and GDELT datasets are illustrated in Figure \ref{fig:neighbor_num}. We select these four graphs because they are relatively dense compared to the others. The sampling number is selected among \{2, 4, 6, 8,10,12,16,32\}. On the UCI dataset, the best performance is attained with a sampling number of 8, with a decline in performance as the sampling number increases. A similar situation is mirrored on the LastFM dataset, where the best performance is achieved with a sampling number of 10. 
On the MOOC dataset, optimal performance is obtained with a large sampling number of 32, whereas the second-best performance occurs with a sampling number of only 2. Similarly, the second-best performance on the LastFM dataset is achieved with a sampling number of 2, demonstrating that a small sampling number can achieve satisfactory results in certain temporal graphs. In contrast, on the GDELT dataset, performance improves as the sampling number increases, reaching its peak at a sampling number of 32.
The results demonstrate that the optimal sampling number varies across different temporal graphs. This demonstrates that in certain cases, a smaller sampling number can achieve near-optimal results while substantially reducing computational overhead. Conversely, there are situations where larger sampling numbers produce superior outcomes.
\subsubsection{Efficiency with GPU Number}

The training time with different numbers of GPUs on the UCI and GDELT datasets is shown in Figure \ref{fig:time_gpu}. Each pair of GPUs is connected via NVLink 400GB/s (P2P NVLink). The training process is accelerated using ZeRO-2 and mixed precision BF16. The training time per epoch on the UCI dataset decreases from 0.79 seconds to 0.25 seconds as the number of GPUs increases from 1 to 8, resulting in a speedup of 3.15$\times$. Similarly, the training time per million edges on the GDELT dataset decreases from 16.53 seconds to 4.54 seconds, achieving a speedup of 3.64$\times$. It can be observed that P2P PCIe exhibits an average performance reduction of approximately 40.65\% compared to P2P NVLink, as concluded in \cite{li2019evaluating,nukada2021performance,choquette2020nvidia}.
These results demonstrate the effectiveness of distributed training in accelerating the process of training TGNNs  with increasing GPU resources.

\section{Related Work}
% TGNNs can be divided into two categories: discrete-time dynamic graphs (DTDG) and continuous-time dynamic graphs (CTDG). DTDG models the temporal graph as a sequence of static graph snapshots taken at discrete time intervals~\cite{pareja2020evolvegcn}, whereas CTDG represents the temporal graph as a stream of node or edge events occurring continuous times~\cite{kumar2019predicting,rossi2020temporal}.
TGNNs incorporate temporal information into graph-based operations to model the evolution of graphs over time. SOTA TGNNs represent the temporal graph as a stream of node or edge-timed events occurring chronologically~\cite{kumar2019predicting,rossi2020temporal}. Pioneering TGNNs incorporate recurrent neural networks (RNNs) to model the embedding trajectories of nodes as the graph evolves~\cite{kumar2019predicting}. The combination of memory modules and graph-based operations makes temporal graph networks (TGNs) a general and efficient framework~\cite{rossi2020temporal}. Temporal graph attention (TGAT) adopts a graph attention layer to efficiently aggregate the features of temporal-topological neighbors and is scalable to relatively large dynamic graphs~\cite{xu2020inductive}. 
The asynchronous propagation attention networks (APAN) decouples the inference and graph computation in TGNNs to accelerate inference efficiency~\cite{wang2021apan}. Motifs captured by causal anonymous walks, are leveraged to improve the performance of TGNNs but suffer from high computational complexity and inefficiency~\cite{paranjape2017motifs, wang2021inductive, jin2022neural}. TGL is the first efficient training framework that integrates neighbor sampling using the CSR representation and model training based on a CPU-GPU mailbox framework implemented with DGL MFGs~\cite{zhou2022tgl}. In addition, ETC is a SOTA generic framework that incorporates novel data batching strategies, enabling larger training batches and reducing redundant data access volume~\cite{gao2024etc}.

\section{Conclusion}
In this paper, we propose TF-TGN, a temporal graph neural network based on the Transformer decoder to model the evolution of temporal graphs. We formulate the temporal message aggregation between chronologically occurring nodes and their temporal neighbors as sequence modeling based on the suffix infilling and temporal graph attention with self-loop operations and unify the training paradigm of TGNNs. We implement TF-TGN using flash-attention and memory-efficient attention mechanisms in a distributed manner, which can greatly speed up the training process, effectively capturing the dynamics in temporal graphs. 
We conduct extensive experiments on 9 real-world temporal graphs with up to billions of edges. Experimental results demonstrate the effectiveness of TF-TGN in terms of prediction accuracy and speedup  compared to existing SOTA TGNN training frameworks.

% We evaluate TF-TGN on 9 real-world temporal graphs with up to billions of edges. The experimental results show that the parallel sampling strategy achieves an average 364.94$\times$ speedup in CSR converting and an average 5.67$\times$ speedup in sampling efficiency. Furthermore,  TF-TGN obtains an average 121.42$\times$ speedup over the classical TGNNs, and even more 2.20$\times$ speedup over the SOTA efficient TGNN training frameworks, while maintaining or even improving prediction accuracy on most datasets.
\bibliographystyle{ACM-Reference-Format}
\bibliography{bib/reference}

\appendix
\section{Datasets}
\label{apd:datasets}
\begin{itemize}
    \item UCI ~\cite{panzarasa2009patterns} is a online community social network containing the messages sent or received by the students at the Universityof California, Irvin.
    \item Wikipedia ~\cite{kumar2019predicting} comprises one month of edits maded by editors on Wikipedia pages. Wikipedia is a bipartite graph with editors and edits as nodes and editing interactions as edges.
    \item Reddit ~\cite{kumar2019predicting} is a bipartite social network representing posts made by users across subreddits within a one-month period.
    \item MOOC ~\cite{kumar2019predicting} is an interaction network illustrating interactions between students and course units within a MOOC platform.
    \item LastFM ~\cite{kumar2019predicting} encompasses one month of interactions detailing which users listen to which songs.
    \item Wiki-Talk ~\cite{paranjape2017motifs, leskovec2010governance} is a temporal network that represents Wikipedia users editing each other's Talk pages. A directed edge in this network indicates that one user has edited another user's talk page at a specific time.
    \item Stack-O. (Stack-Overflow) ~\cite{paranjape2017motifs} is a temporal network that captures interactions on the Stack Exchange website, Stack Overflow. Each interaction in this network signifies a user either answering or commenting on another user's question or answer.
    \item GDELT ~\cite{zhou2022tgl} is a temporal knowledge graph derived from the Event Database within GDELT 2.0 ~\cite{Leetaru13gdelt:global}, documenting global events sourced from news and articles in over 100 languages, updated every 15 minutes.
    \item MAG ~\cite{zhou2022tgl} is a subset of the heterogeneous MAG240M graph within OGB-LSC ~\cite{hu2021ogb}, focusing on a paper-paper citation network. Each node represents an academic paper, with directional temporal edges indicating citations from one paper to another, timestamped by the publication year of the citing paper.
\end{itemize}

\section{Baselines}
\label{apd:baselines}
\begin{itemize}
    \item TGAT~\cite{xu2020inductive}, which stands for Temporal Graph Attention Networks, utilizes self-attention to aggregate temporal-topological neighborhood features and captures useful time-event interactions through a functional time encoding approach.
    \item TGN~\cite{rossi2020temporal}, abbreviated as Temporal Graph Networks, is a generic and efficient framework for learning temporal graph representations. It models the temporal evolution of graph structure and node features, incorporating a memory module and graph-based operators.
    \item APAN~\cite{wang2021apan} is an Asynchronous Propagation Attention Network that aims to decouple model inference from graph computation to mitigate the impact of intensive graph query operations on model inference speed.
    \item TGL~\cite{zhou2022tgl} is a general framework that integrates neighbor sampling using the CSR representation and model training based on a CPU-GPU mailbox framework implemented with DGL MFGs, aimed at efficiently training large-scale dynamic graphs.
    \item ETC~\cite{gao2024etc} is a generic framework that incorporates novel data batching strategies, enabling larger training batches and reducing redundant data access volume, thereby improving model computation efficiency and achieving significant training speedup.
\end{itemize}

\section{Sampling Algorithm}
\label{apd:sampling}
The space complexity of Algorithm \ref{alg:sampling} is $\mathcal{O}(|\mathcal{V}|+2|\mathcal{E}|)$. The computational complexity of converting the temporal interactions of a dynamic graph into the T-CSR representation is $\mathcal{O}(n + \frac{n}{\bar{k}}\bar{k}\log{\bar{k}})$ = $\mathcal{O}(n + n\log{\bar{k}})$.  The computational complexity of sampling the temporal neighbors of each node is $\mathcal{O}(1 + \log{\bar{k}})$, where $n$ is the number of temporal edges $|\mathcal{E}|$, $\bar{k}$ is the average number of temporal neighbors of each node. The sampling algorithm implemented in parallel can significantly reduce the practical running time and accelerate the training process on large-scale temporal graphs. In the \textit{sample} function in Algorithm \ref{alg:sampling}, we provided two sampling strategies: 1) randomly sample $k$ temporal neighbors of each node; 2) sample the top-$k$ recent temporal neighbors of each node. Algorithm \ref{alg:sampling} can be easily extended to other sampling strategies and can be integrated into the distributed training process.

Moreover, Algorithm \ref{alg:sampling} also works for static graphs, where the temporal interactions of a dynamic graph can be converted to the CSR representation of the static graph without sorting the timestamps of the edges. The computational complexity of converting the temporal interactions of a dynamic graph to the CSR representation is $\mathcal{O}(n)$. The computational complexity of sampling the neighbors of each node is $\mathcal{O}(1)$.

\end{document}